\documentclass[10pt,letterpaper]{article}

\usepackage[pagenumbers]{cvpr} 

\usepackage{amsmath}
\usepackage{amssymb}
\usepackage{booktabs}
\usepackage{array}  
\usepackage{algorithmic}
\usepackage{textcomp}
\usepackage{xcolor}
\usepackage{url}
\usepackage{latexsym}
\usepackage{graphicx, epsfig} 
\usepackage{longtable}
\usepackage{subcaption}
\usepackage{hyperref}
\hypersetup{
    colorlinks=true,
    linkcolor=blue!50!cyan, 
    filecolor=magenta,      
    urlcolor=blue!50!cyan, 
    citecolor=blue!50!cyan, 
}

\usepackage[round,semicolon]{natbib}
\usepackage{fancyhdr} 

\usepackage[utf8]{inputenc}
\usepackage{lineno}  

\usepackage[capitalize]{cleveref}
\crefname{section}{Sec.}{Secs.}
\Crefname{section}{Section}{Sections}
\Crefname{table}{Table}{Tables}
\crefname{table}{Tab.}{Tabs.}

\newcommand\keywords[1]{\textbf{Keywords}: #1}

\def\cvprPaperID{*****} 
\def\confName{CVPR}
\def\confYear{2023}

\begin{document}


\title{Generative AI Models for Different Steps in Architectural Design: A Literature Review}

\author{
  Chengyuan Li$^{a}$ \quad Tianyu Zhang$^{b}$ \quad Xusheng Du$^{b}$ \quad Ye Zhang$^{a}$\thanks{
Corresponding author at: School of Architecture, Tianjin University, Tianjin (300000), China. E-mail address: yaapp2012@gmail.com (Ye Zhang)
  } \quad Haoran Xie$^{b}$ 
  \\
  \vspace{0.3em} \\
  {$^a$School of Architecture, Tianjin University, Tianjin (300000), China} \quad \\
  \makebox[\textwidth]{\parbox{0.95\textwidth}{\centering $^b$Graduate School of Advanced Science and Technology, Japan Advanced Institute of Science and Technology, Ishikawa (923-1292), Japan}}
}

\maketitle
\begin{abstract}

Recent advances in generative artificial intelligence (AI) technologies have been significantly driven by models such as generative adversarial networks (GANs), variational autoencoders (VAEs), and denoising diffusion probabilistic models (DDPMs). Although architects recognize the potential of generative AI in design, personal barriers often restrict their access to the latest technological developments, thereby causing the application of generative AI in architectural design to lag behind. Therefore, it is essential to comprehend the principles and advancements of generative AI models and analyze their relevance in architecture applications. This paper first provides an overview of generative AI technologies, with a focus on probabilistic diffusion models (DDPMs), 3D generative models, and foundation models, highlighting their recent developments and main application scenarios. Then, the paper explains how the abovementioned models could be utilized in architecture. We subdivide the architectural design process into six steps and review related research projects in each step from 2020 to the present. Lastly, this paper discusses potential future directions for applying generative AI in the architectural design steps. This research can help architects quickly understand the development and latest progress of generative AI and contribute to the further development of intelligent architecture.

\end{abstract}

\noindent

\section{Introduction}

\begin{figure}[t]
    \centering
    \includegraphics[width=0.85\linewidth]{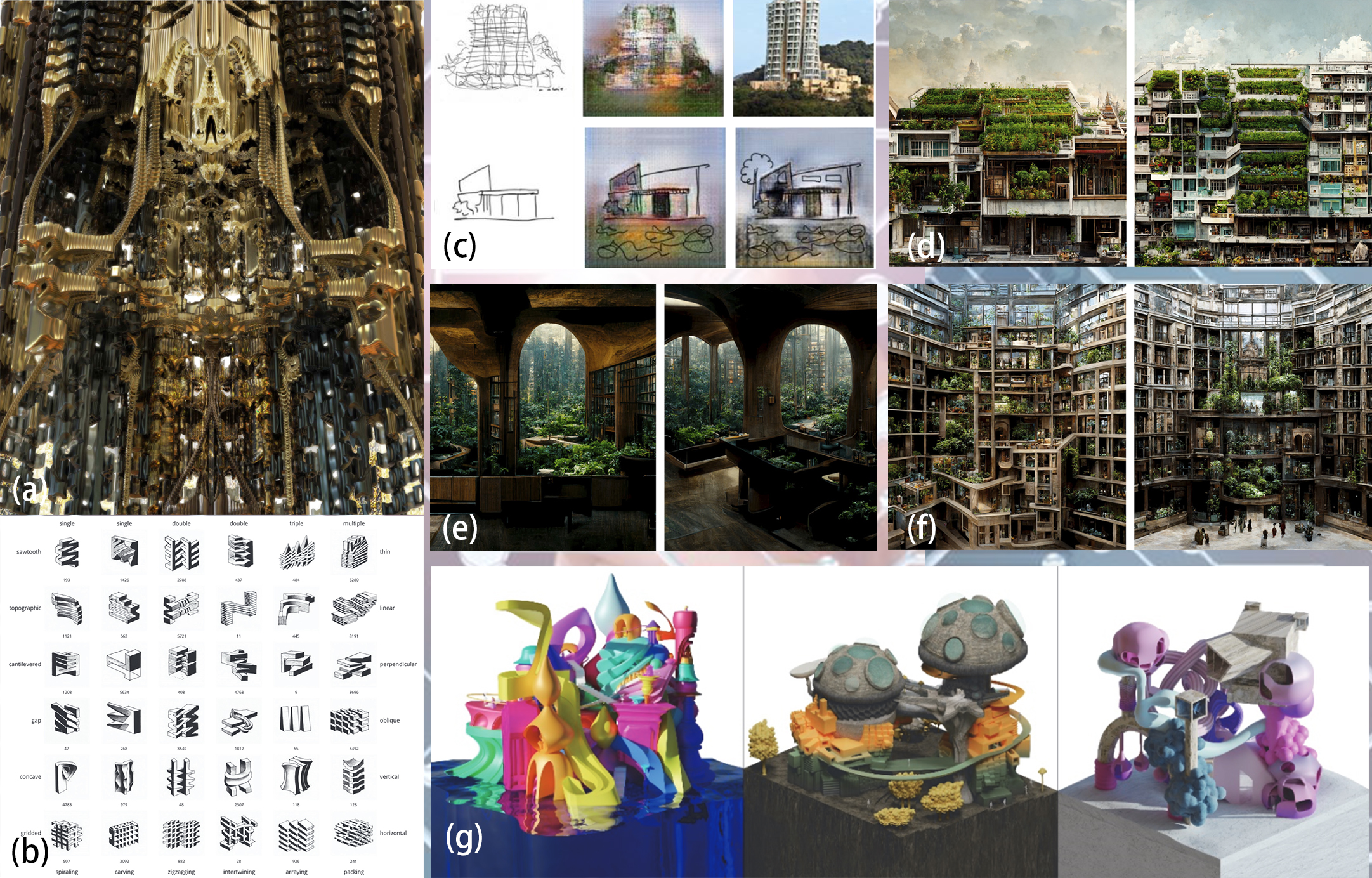}
    \caption{Examples of architecture design using generative AI techniques: (a) church design~\citep{del2019church}; (b) matrix of cuboid shapes~\citep{kim2023text2form}; (c) Frank Gehry’s Walt Disney concert hall~\citep{zhuang1973rendering}; (d) Bangkok urban design~\citep{koehler2023more}; (e) foresting architecture~\citep{koehler2023more}; (f) urban interiors ~\citep{koehler2023more}; (g) text-to-architectural design~\citep{stigsenai}.}
    \label{fig:genai}
\end{figure}

Generative artificial intelligence (AI) technologies—which create diverse content such as text, images, music, videos, and 3D models—are rapidly advancing and reshaping architectural design. Traditional generation AI models such as generative adversarial networks (GANs) and variational autoencoders (VAEs) have been extensively utilized in architectural image generation, as illustrated in Figure \ref{fig:genai}. With the development of large visual models —Stable Diffusion, DALL-E 2, and Midjourney— architects and researchers are increasingly using generative AI for creative design generation ~\citep{zhang2023exploringthe,tan2024using,sukkar2024analytical,horvath2024ai,baudoux2024benefits} 
and to assist in the architectural design process ~\citep{zhong2024ai,del2022can,he2022sequential,newton2019generative}. 
In architectural design practice, Metropolitan Architecture Research Unit Van Rijs De Vries (MVRDV) and Zaha Hadid Architects use generative AI tools to improve design efficiency by generating conceptual architectural images. Architects also use generative AI tools in their work; the Royal Institute of British Architects (RIBA) reports that 41\% of UK architects have occasionally used AI in projects, with 43\% believing that it enhances the efficiency of the design process ~\citep{riba2024ai}.

\begin{figure}[t]
    \centering
    \includegraphics[width=0.95\linewidth]{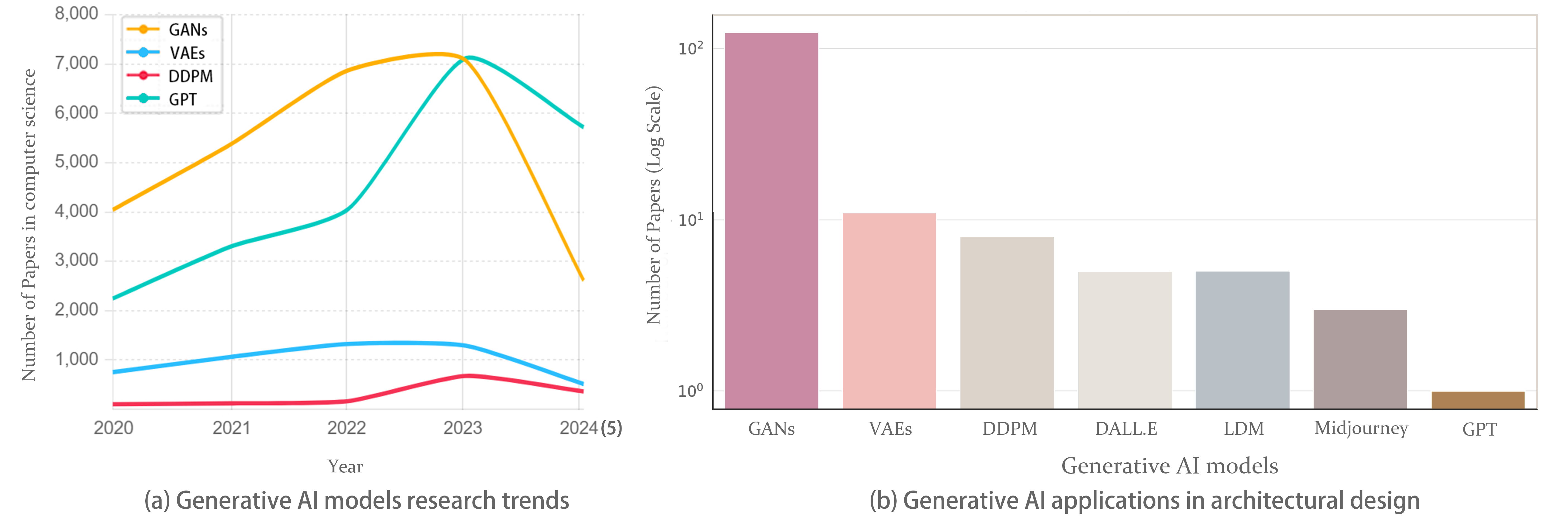}
    \caption{Parallel analysis of overall research trends in generative AI and applications in architectural design.}
    \label{1}
\end{figure}

However, the application of generative AI in architectural design continues to face an apparent lag.
As shown in Figure~\ref{1}(a), in computer science, GANs and VAEs are the most commonly used generative AI models. The number of research papers on denoising diffusion probabilistic models (DDPMs), latent diffusion models (LDMs), and generative pre-trained transformers (GPTs) is increasing. In contrast, as illustrated in Figure~\ref{1}(b), the application of these advanced models in architectural design remains relatively limited compared to GANs and VAEs. 
This indicates a significant lag in adopting new generative AI models within the field of architecture, possibly because of personal barriers \citep{sourek2024ai}, as the complexity of AI algorithms and models that require extensive expertise. Architects find it challenging to adopt new technologies quickly, thereby prompting them to opt for traditional design methods.
Against this background, the research objective of this paper is to delineate the application of generative AI in different architecture design steps. 
This research is based on reviews of randomly sampled journals and conferences, including related journals in architectural design, engineering, urban planning, and computer science, as well as related conferences in architectural design, architectural engineering, and computer science.  

\begin{figure}[t]
    \centering
    \includegraphics[width=0.85\linewidth]{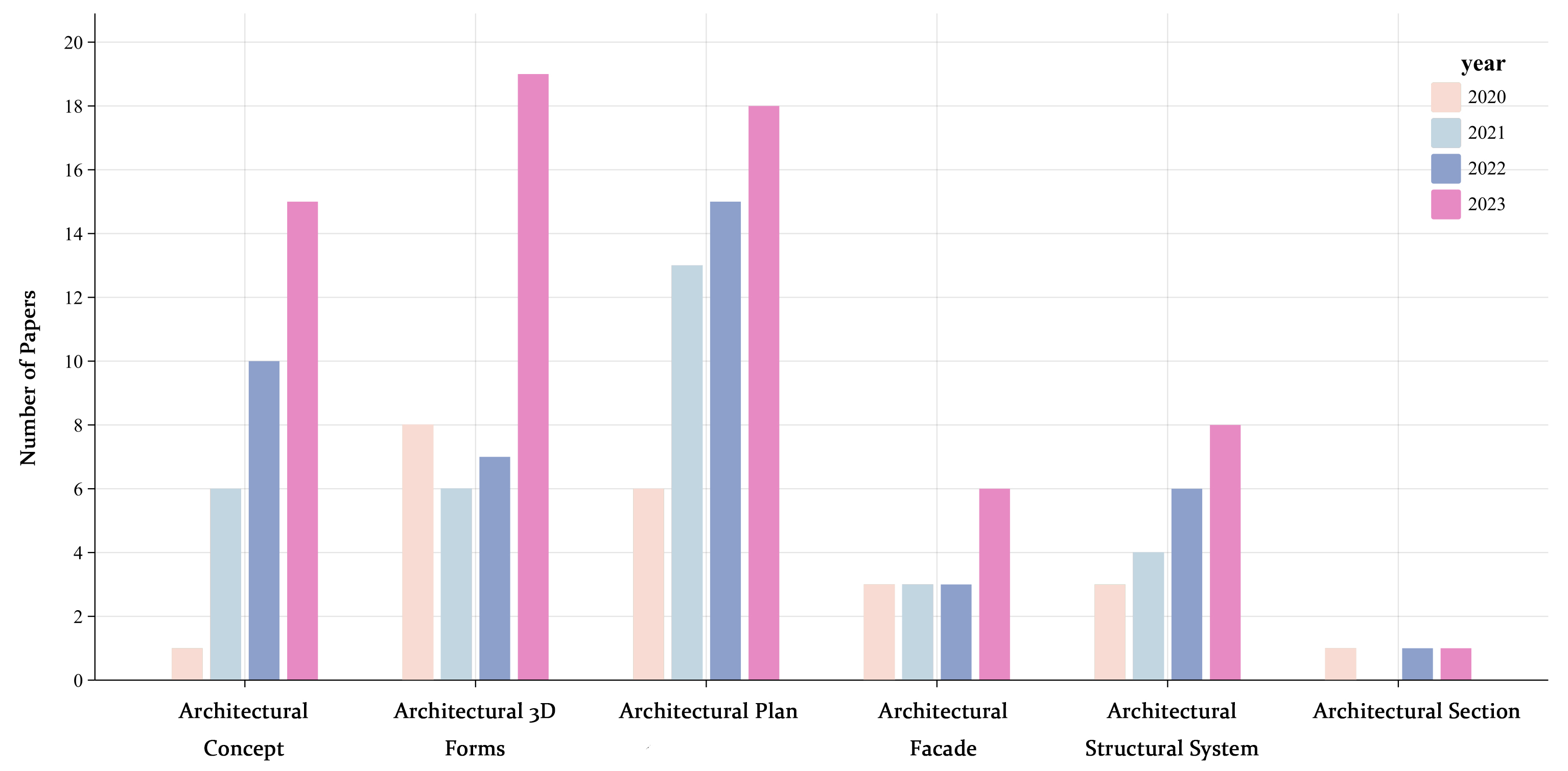}
    \caption{The number of research papers using generative AI technology to various architectural design steps.}
    \label{2}
\end{figure}

To clearly illustrate the application of AI in various architectural design steps, this paper outlines a detailed breakdown of these steps.
The architectural design process moves from abstraction to concrete realization, following logical steps. 
First, the overall design direction is established through a design concept. This is then translated into a tangible form during the 3D form generation step, which defines the building's space and volume. Next, the floor plan refines the functional layout and internal spatial relationships, followed by the structural system, which ensures the building's safety and physical feasibility. Facade design follows, showcasing the building's aesthetic and harmony with the surrounding environment. Finally, section design is derived naturally, synthesizing the spatial and structural elements to complete the architectural representation. Each step is interconnected, building progressively to form a cohesive design system.
The design steps described above are not directly outlined in research literature but is summarized from "The Professional Practice of Architectural Working Drawings" ~\citep{wakita2003professional}. The specific design steps in book vary based on a project's function and type. All projects include key steps such as floor plans, elevations, sections, and structural drawings. Although conceptual design varies in detail and presentation, it is common in most projects. In some cases, a 3D architectural model is also incorporated. The integrated design steps across these projects align well with existing studies, which focus on applying generative AI in specific architectural design steps, as illustrated in Figure~\ref{2}. 
Based on the context above, this paper categorizes how generative AI is applied in the architectural design process into the following six steps: concept image design, architectural 3D form design, floor plan design, facade design, structural system design, and section design. 

The remainder of this article is structured in the following manner: Section 2 provides a detailed introduction to various generative AI models—focusing on GANs, diffusion models (DMs), 3D generative models, and foundation models—and discusses the applications of these models in generating images, videos, and 3D models. Section 3 explores the application of generative AI in various architectural design tasks, comparing input and output data types and generation models. It also categorizes these applications to distinguish different scenarios. Section 4 explores the potential applications of generative AI models in the future.

\section{Generative AI Models}
\label{sec:gam}

\begin{figure*}[t]
    \centering
    \includegraphics[width=0.95\linewidth]{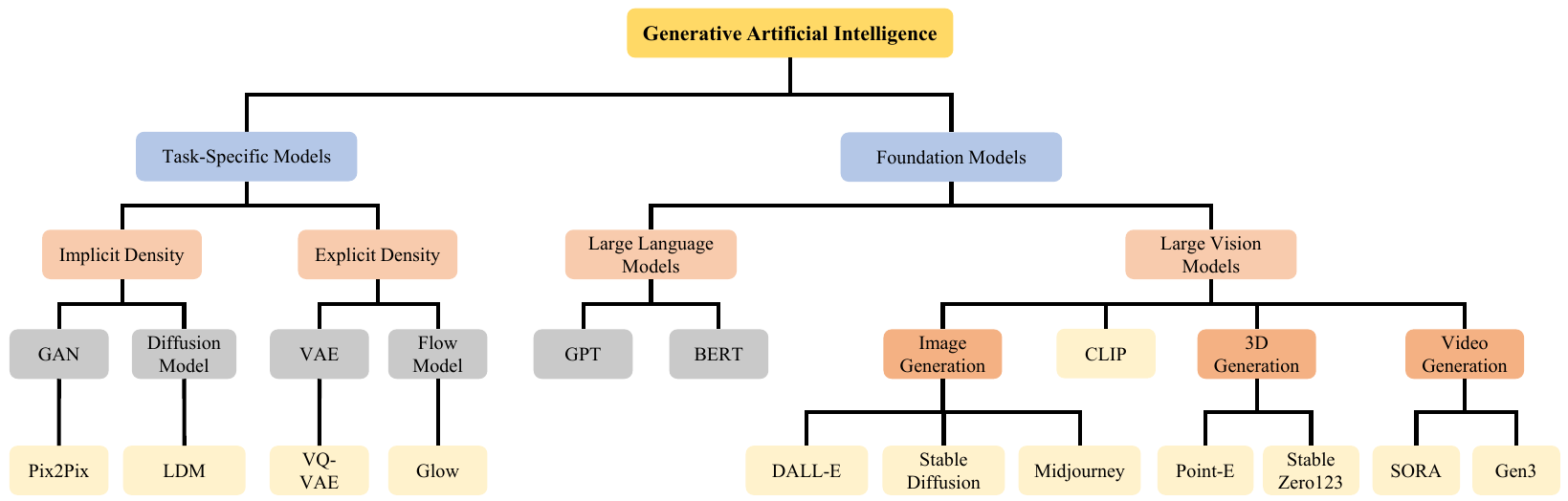}
    \caption{Types of generative artificial intelligence models.}
    \label{fig:GAI}
\end{figure*}

Generative AI models are rapidly advancing with the continual emergence of new methods, as illustrated in Figure~\ref{fig:GAI}. Deep learning-based approaches, including VAEs, GANs, DMs, flow models, and transformer models have significantly enhanced image and text generation techniques. VAEs were pioneering generative models, using an encoder–decoder architecture with probabilistic graphical models to learn latent representations for image generation \citep{gregor2015draw}. GANs marked a milestone by employing a generator and discriminator in an adversarial training process, driving the generator to create images that increasingly resembled real data \citep{karras2019style, yi2019apdrawinggan}. DMs have recently emerged as revolutionary, demonstrating outstanding image generation quality \citep{rombach2022high,ramesh2021zero}. Flow models achieve precise probability estimation and efficient sample generation through reversible transformations \citep{papamakarios2021normalizing}. In addition, transformer models, with their self-attention mechanism, have significantly improved sequential data processing and form the basis of many generative models \citep{vaswani2023attentionneed}. In this paper, we focus on the widely used GANs in architecture and the state-of-the-art Diffusion Models.

Large language models (LLMs)—like the generative pre-trained transformer (GPT) \citep{radford2018improving} series and bidirectional encoder representations from transformers (BERT) \citep{devlin2018bert}, based on transformer architecture—are trained on vast text data, thereby revealing strong language understanding and generation capabilities that can be applied in tasks such as text generation, answering questions, and machine translation. Large vision models use deep learning techniques for image data processing and generation, typically based on convolutional neural networks (CNNs) and transformer architectures, excelling in image classification, object detection, and image generation. These large models have gained significant attention in academia and practical applications, pushing the boundaries of image and text generation technologies and highlighting the potential and broad application of deep learning in generative models. 

\subsection{Generative Adversarial Networks}

\begin{figure}[t]
    \centering
    \includegraphics[width=0.95\linewidth]{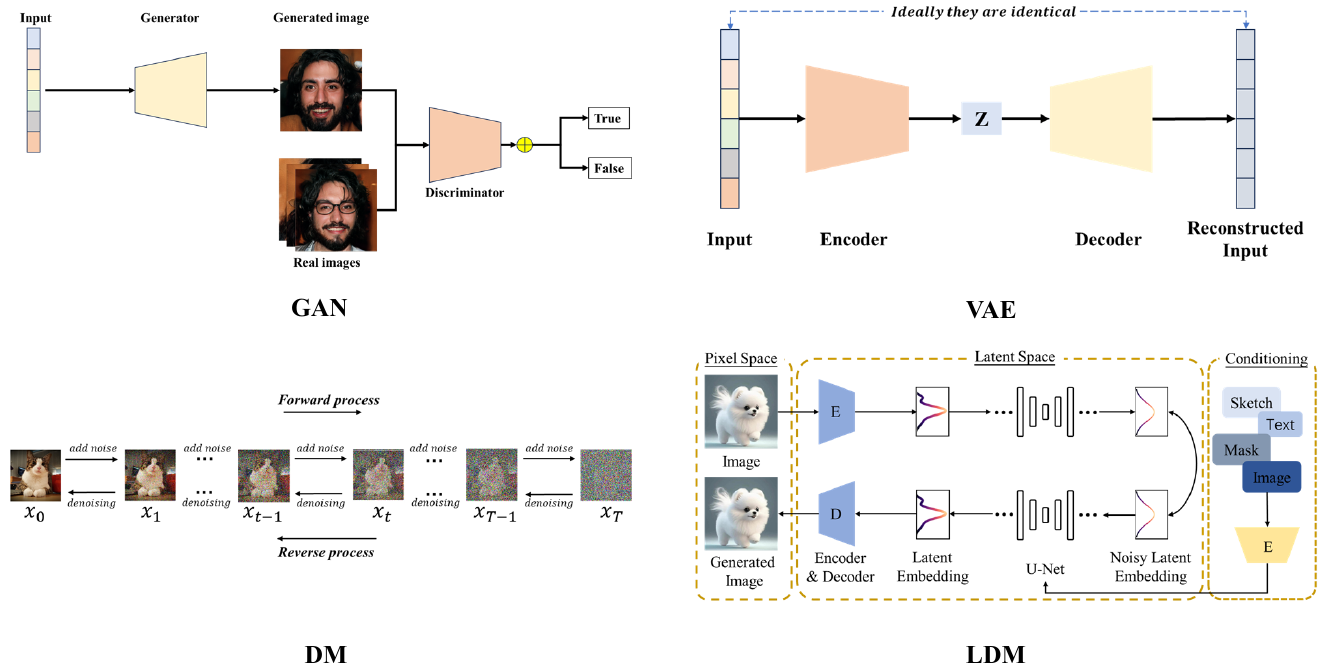}
    \caption{The framework of Generative Adversarial Networks (GAN), Variational Autoencoders (VAE), Diffusion Models (DM) and Latent Diffusion Models (LDM). Where $z$ is a compressed low-dimensional representation of the input.}
    \label{fig:GAN}
\end{figure}

GANs \citep{goodfellow2014generative} comprise a generator and a discriminator, as illustrated in Figure~\ref{fig:GAN}. The generator produces synthetic samples from random noise, while the discriminator evaluates whether these synthetic samples resemble real samples. The objective is for the generator to create samples indistinguishable from real data according to the discriminator. This adversarial framework enables the model to maintain a dynamic equilibrium between generation and discrimination, thereby driving the learning and optimization of the entire system. Despite the many advantages of GANs, they also face challenges, such as mode collapse during training. This occurs when the generator overly focuses on a few patterns or samples, thereby neglecting the broader diversity of the dataset and resulting in outputs that are limited, repetitive, or lack diversity. This issue impacts the robustness and generalization of the model.

\paragraph{Conditional GAN:}

To address the limited controllability of traditional GAN models, Conditional GAN (CGAN) \citep{mirza2014conditional} was introduced. CGAN controls the image generation process by incorporating conditional informa- tion—such as text, labels, or hand-drawn sketches—to produce images that meet specific criteria. In CGAN, the generator receives both random noise and conditional information, thereby enabling it to produce images that align more closely with the given conditions. This approach enhances the precision of the generated results. Additionally, variants such as pix2pix \citep{isola2017image} and StyleGAN \citep{karras2019style} have been developed to further refine and extend the capabilities of conditional image generation.

\subsection{Diffusion Models}

In image generation, DMs outperform GANs and VAEs \citep{song2019generative,ho2020denoising}. 
Most DMs that are currently used are based on denoising diffusion probabilistic model (DDPM) \citep{ho2020denoising}, which simplifies the DM through variational inference.  
As shown in Figure~\ref{fig:GAN}, DMs include both forward diffusion and reverse denoising (inference) processes. The forward process follows the concept of a Markov chain and turns the input image into Gaussian noise. Given a data sample $x_0$, the Gaussian noise is progressively increased to the data sample during $T$ steps in the forward process, producing noisy samples $x_t$. As the timestep increases, the distinguishable features of $x_0$ gradually diminish. Eventually When $T$ tends to infinity, $x_T$ is equivalent to a Gaussian distribution with isotropic covariance. In addition, the inference process can be understood as a sequence of denoising autoencoders with the same weights (autoencoder is typically implemented as U-Net \citep{ronneberger2015u}, which are trained to forecast denoised images of their corresponding inputs, $x_t$. 

\paragraph{Latent Diffusion Model:}

Different from DDPM, LDM \citep{rombach2022high} do not directly operate on the images but operate in the latent space called perceptual compression. LDMs reduce the dimensionality of the data by projecting it into an efficient, low-dimensional latent space in which imperceptible high-frequency details are abstracted away. The framework of an LDM is illustrated in Figure~\ref{fig:GAN}. After the image is compressed by the encoder to latent representation, the diffusion process is performed in the latent representation space. Finally, LDM infers the data sample from the noise and the decoder restores the data to the original pixel space and obtains the generated images. To accelerate the generation speed, the latent consistency model (LCM) \citep{luo2023latent} was proposed to optimize the step of denoising inference.

\subsection{Foundation Models}

\begin{figure*}[t]
    \centering
    \includegraphics[width=0.95\linewidth]{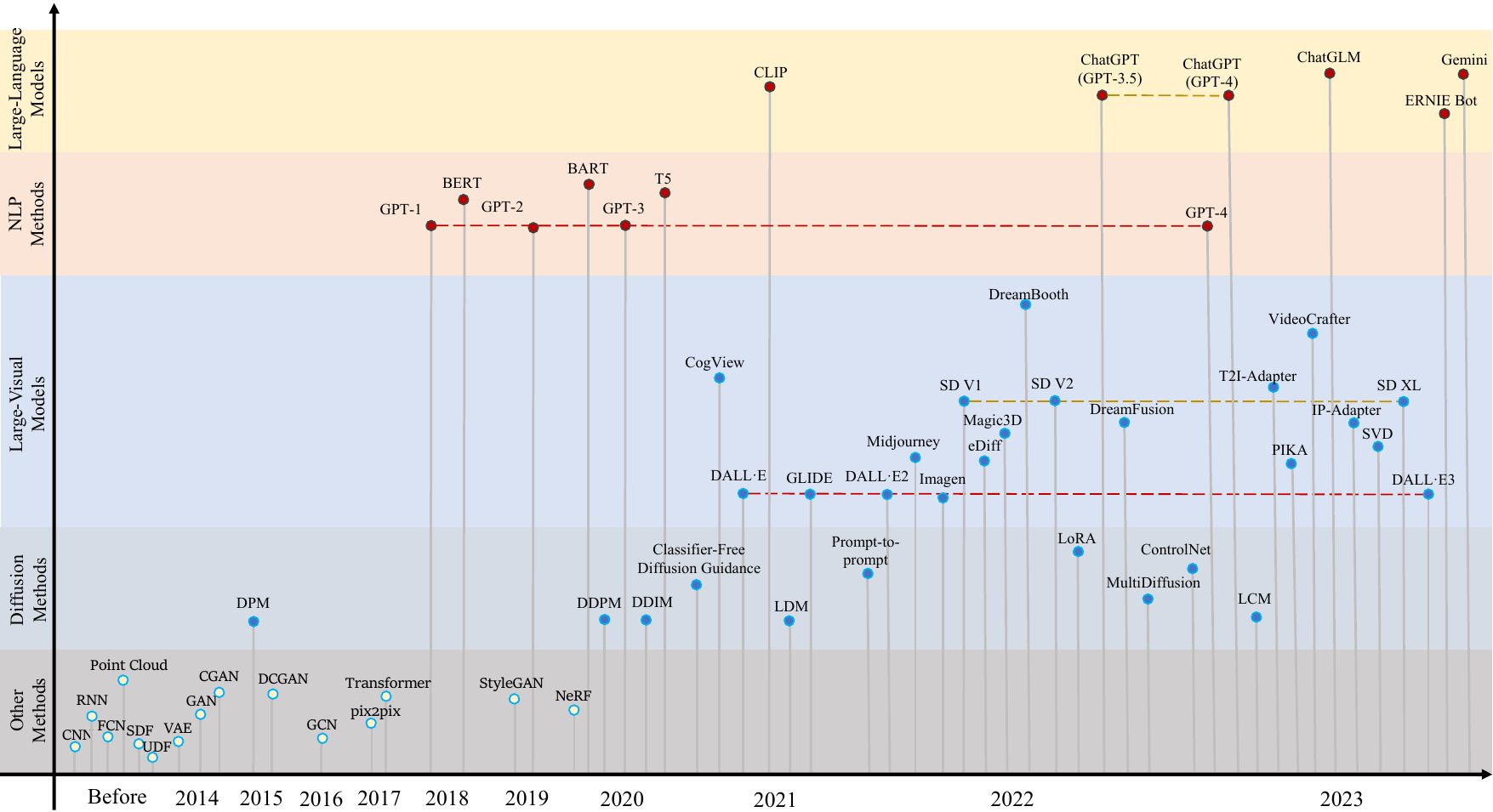}
    \caption{The evolution of prominent large-scale models in computer science.}
    \label{fig:LSM}
\end{figure*}

In computer science, foundation models, also called large-scale models use deep learning models with numerous parameters and intricate structures, particularly in uageguage processing and computer vision tasks. These models demand substantial computational resources for training but exhibit exceptional performance across diverse tasks. The evolution from basic neural networks to sophisticated DMs, as depicted in Figure~\ref{fig:LSM}, illustrates the continuous quest for more robust and adaptable AI systems.

\subsubsection{Large Language Models (LLM)}
\label{LLM}

The transformer model has achieved remarkable success in natural language processing (NLP) which consists of several components: encoder, decoder, positional encoding, and the final linear and SoftMax layers. Both the encoder and decoder comprise multiple identical layers. Each layer contains several components of attention layers and feedforward network layers. Additionally, positional encoding is used to inject positional information into the text embeddings, thereby indicating the position of words within the sequence. 
Notably, transformer has paved the way for two prominent transformer models: BERT \citep{devlin2018bert} and GPT \citep{radford2018improving}. 
The main difference is that BERT is based on a bidirectional pre-training language model and fine-tuning, while GPT is based on an autoregressive pre-training language model and prompting. 

GPT aims to pre-train models using large-scale unsupervised learning to facilitate the generation and understanding of natural language. The training process involves two primary stages: Initially, a language model is trained in an unsupervised manner on extensive corpora without task-specific labels or annotations. Subsequently, supervised fine-tuning occurs during the second stage, catering to specific application domains and tasks. In addition, BERT has emerged as a breakthrough approach, achieving state-of-the-art performance across diverse language tasks. BERT’s training methodology comprises two key stages: pre-training and fine-tuning. Pre-training involves the utilization of extensive text corpora to train the language model. 
The primary objective of pre-training is to endow the BERT model with robust language understanding capabilities, thereby enabling it to effectively tackle various NLP tasks. Subsequently, fine-tuning utilizes the pre-trained BERT model in conjunction with smaller labeled datasets to refine the model parameters. This process facilitates the customization of the model to specific tasks, thereby enhancing its suitability and performance for targeted applications, such as sentiment analysis and text classification. 

In recent years, LLMs have witnessed rapid and explosive growth. Basic language models refer to models that are only pre-trained on large-scale text corpora, without any fine-tuning. Examples of such models include the language model for dialogue applications (LaMDA) \citep{thoppilan2022lamda} and OpenAI's GPT-3 \citep{brown2020language}. 

\subsubsection{Large Vision Models}

In computer vision, pretrained vision-language models such as contrastive language-image pre-training (CLIP) \citep{radford2021learning} have demonstrated powerful zero-shot generalization performance across various downstream visual tasks. These models are typically trained on hundreds of millions to billions of image–text pairs collected from the web. In addition, certain research efforts also focus on large-scale base models conditioned on visual input prompts. For example, the segment anything model (SAM)  \citep{kirillov2023segment} can perform category-agnostic segmentation from given images and visual prompts (such as boxes, points, or masks).

The current generative models based on DMs present unprecedented creative and understanding capabilities. Stable diffusion \citep{rombach2022high} uses the CLIP \citep{radford2021learning} text encoder and can adjust the model through text prompts. Its diffusion process begins with random noise and gradually denoising until a complete data sample is generated. DALLE-3 \citep{ramesh2022hierarchical} utilizes DMs with massive data to generate amazing results. Midjourney excels at adapting to actual artistic styles to create images with any combination of effects the user desires.

\begin{figure*}[t]
    \centering
    \includegraphics[width=0.98\textwidth]{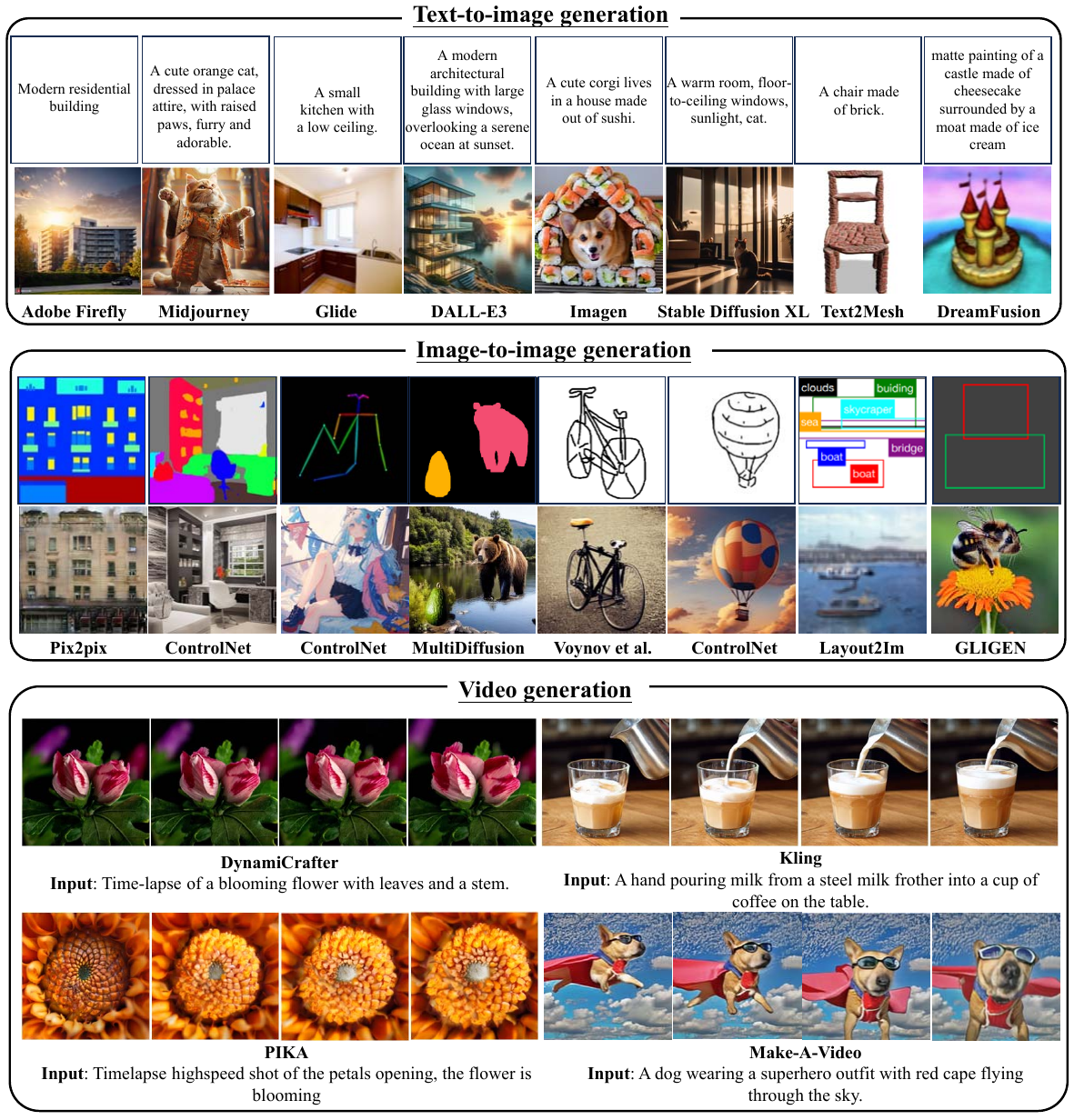}
    \caption{Some examples of generated results from image generation and video generation models.}
    \label{fig:gm_r}
\end{figure*}

\subsection{Applications of 2D Generative AI}

In this section, we introduce widely used applications of generative AI, including image generation (Section~\ref{sec:ig}) and video generation (Section~\ref{sec:vg}). Furthermore, we present results from presented models in Figure~\ref{fig:gm_r} as illustrative references.

\subsubsection{Image Generation}
\label{sec:ig}

\paragraph{Text-to-image:}

Synthesizing high-quality images from text descriptions is a challenging problem in computer vision, StackGAN \citep{zhang2017stackgan} proposed a two-stage model to solve this issue. In the first stage, StackGAN generates the primitive shape and colors of the object based on the given text description, thus yielding initial low-resolution images. In the second stage, StackGAN takes the low-resolution result and text prompts as inputs and generates high-resolution images with photo- realistic details. It can rectify defects in the results of the first stage and add exhaustive details during the refinement process. Guided language to image diffusion for generation and editing (GLIDE) \citep{nichol2021glide} extends the core concepts of DMs by adding additional text information to enhance the training process, ultimately generating text- conditioned images. Based on these foundations—utilizing DMs and extensive data—with the release of LDM \citep{rombach2022high}, stable diffusion based on LDM has also sprung up. These works  cover areas such as image editing and more powerful 3D generation, further advancing image generation and making it closer to human needs.

\paragraph{Image-to-image:}

Image domain refers to a specific category or style of images characterized by distinct visual attributes, such as color, texture, or semantic content. Image-to-image translation can convert the content in an image from one image domain to another with cross-domain conversion between images. The objective of sketch-to-image generation is to ensure that the generated image maintains consistency in both context and appearance with the provided hand-drawn sketch. Pix2Pix \citep{isola2017image} stands out as a classic GAN model capable of handling diverse image translation tasks, including the transformation of sketches into fully realized images. In addition, SketchyGAN \citep{chen2018sketchygan} focuses on the sketch- to-image generation task and aims to achieve more diversity and realism. Currently, ControlNet \citep{zhang2023adding} can control DMs by adding extra conditions, such as sketch, layout and masks. The sketch-to-image generation tasks are applied in both photo-realistic and anime-cartoon styles \citep{peng2023difffacesketch,anifacedrawing}. The layout typically encompasses details such as the position, size, and relative relationships of individual objects. Layout2Im \citep{Zhao_2019_CVPR} is designed to take a coarse spatial layout, consisting of bounding boxes and object categories, for generating a set of realistic images. These images accurately depict the specified objects in their intended locations. 
To enhance the global attention in context, \cite{He_2021_CVPR} introduced the context feature conversion module to ensure that the generated feature encoding for objects remains aware of other coexisting objects in the scene. With regard to DMs, GLIGEN \citep{li2023gligen} facilitates grounded text-to-image generation using prompts and bounding boxes as condition inputs in open worlds. 

\subsubsection{Video Generation}
\label{sec:vg}

Since text prompts only generate some discrete tokens, text-to-video generation is more difficult than tasks such as image retrieval and image captioning. The video diffusion model \citep{ho2022video} is the first to use a DM for video generation tasks. The video DM proposes 3D UNet, which can be applied on variable sequence lengths. Thus, it can be jointly trained on video and image modeling goals, thereby making it suitable for video generation tasks. Additionally, Make-A-Video \citep{singer2022make} is based on the pre-trained text-to-image model and adds one-dimensional convolution and attention layers in the time dimension to transform it into a text2video model. By learning the connection between text and vision through the T2I model, the single-modal video data is utilized to learn the generation of temporal dynamic content. Furthermore, the consistency and controllability of video generation models have also garnered increased attention from researchers.
PIKA \citep{wang2023pika} has been proposed to support dynamic transformations of elements in a scene  based on prompts, without causing the overall image to collapse. DynamiCrafter \citep{xing2023dynamicrafter} utilizes pre-trained video diffusion priors to add animation effects to static images based on textual prompts. This tool supports high-resolution models and, thus, provides better dynamic effects, higher resolution, and stronger consistency. Recently, Kling has brought impressive video generation quality and effects, excelling in aspects such as detailed camera work, lighting variations, adherence to physics, and aesthetic appeal.

\subsection{3D Generative Models}
In addition to 2D images, 3D models have a wide range of applications in architecture. In this section, we introduce various methods and representations used in computer graphics and vision to generate 3D models. We also discuss advances in text-to-3D and image-to-3D modeling techniques.

\subsubsection{3D Shape Representation}

\begin{figure}[t]
    \centering
    \begin{subfigure}{0.15\linewidth}
        \includegraphics[width=\linewidth]{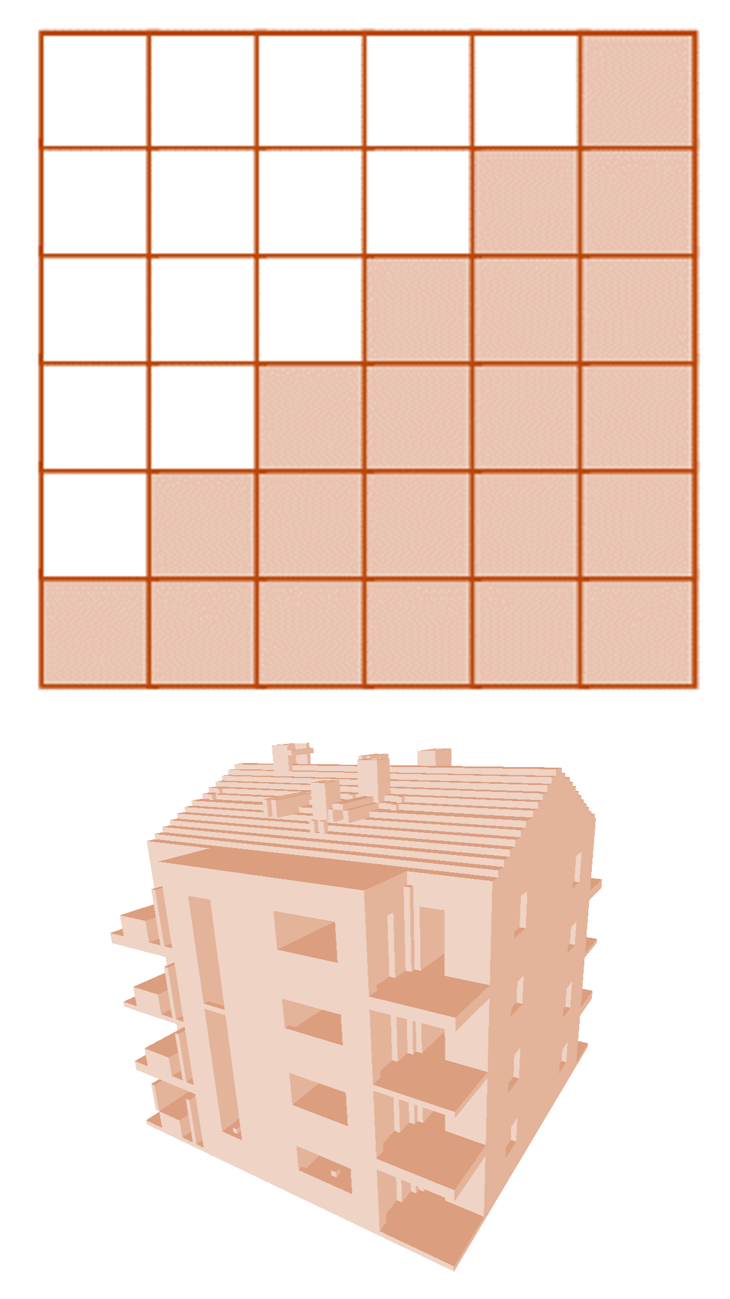}
        \caption{Voxel}
        \label{fig:rp_a}
    \end{subfigure}
    \hspace{0.05\linewidth}
    \begin{subfigure}{0.15\linewidth}
        \includegraphics[width=\linewidth]{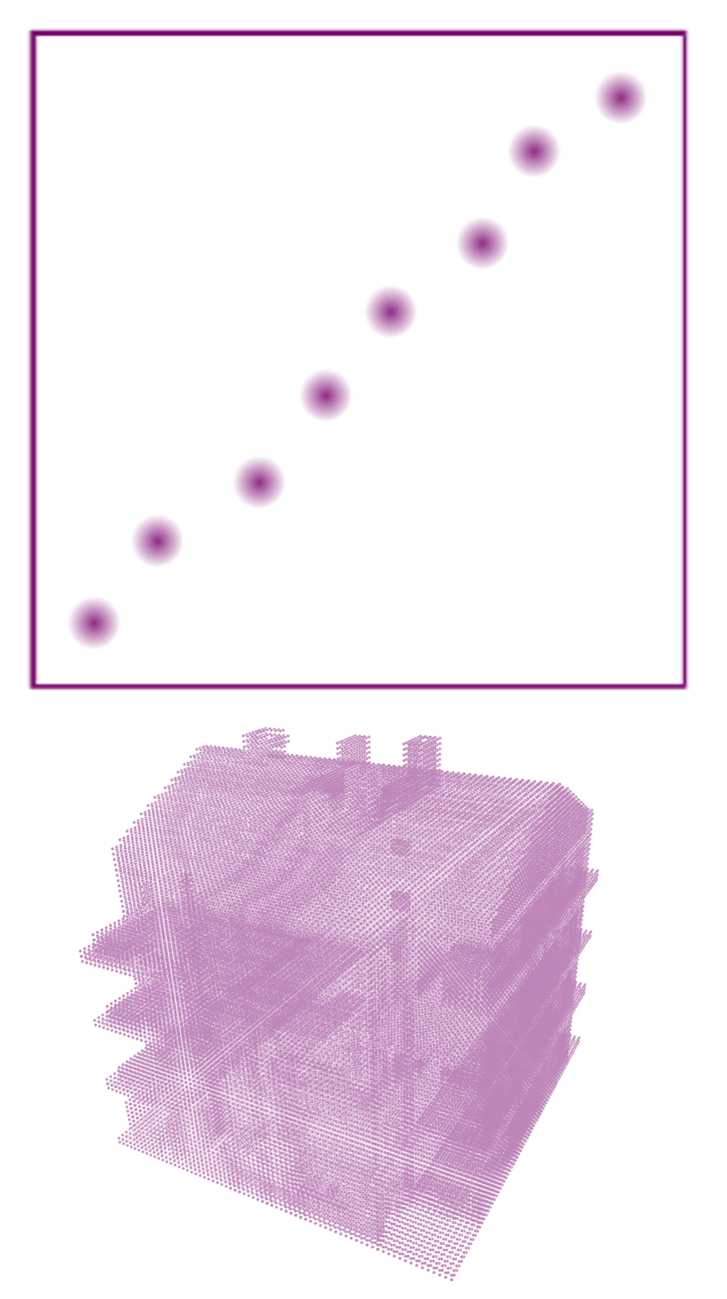}
        \caption{Point}
        \label{fig:rp_b}
    \end{subfigure}
    \hspace{0.05\linewidth}
    \begin{subfigure}{0.15\linewidth}
        \includegraphics[width=\linewidth]{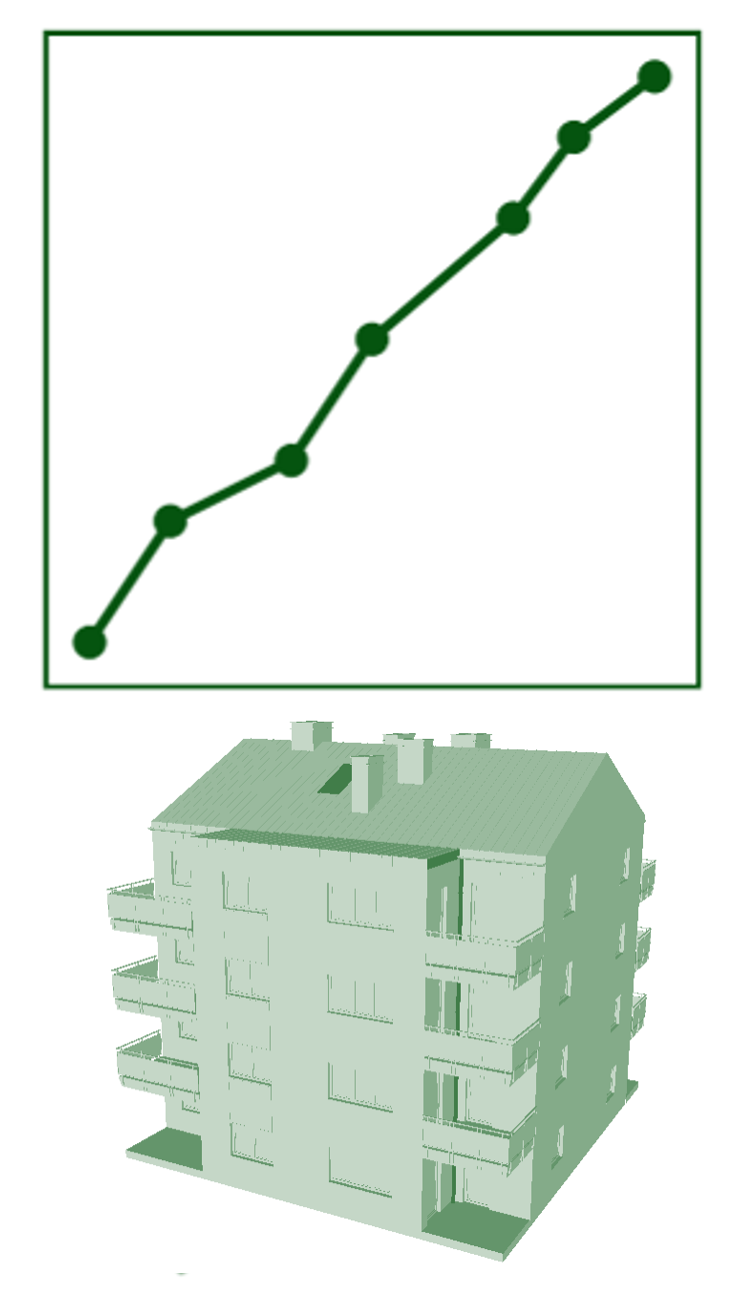}
        \caption{Mesh}
        \label{fig:rp_c}
    \end{subfigure}
    \hspace{0.05\linewidth}
    \begin{subfigure}{0.15\linewidth}
        \includegraphics[width=\linewidth]{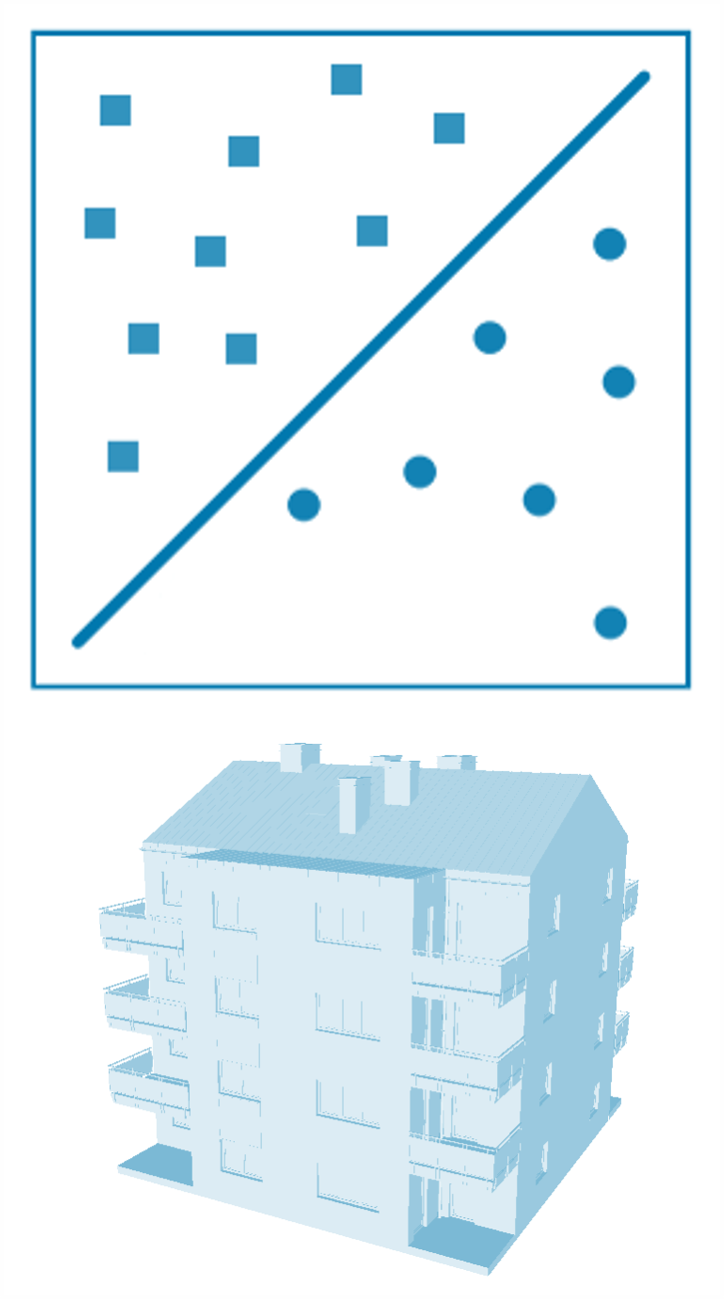}
        \caption{Implicit}
        \label{fig:rp_d}
    \end{subfigure}
    \caption{Representation examples of 3D shapes.}
    \label{fig:rp}
\end{figure}

Representation in 3D visual problems can generally be divided into four categories: voxel-based, point cloud-based, mesh- based, and implicit representation-based. 
As shown in Figure~\ref{fig:rp_a}, the voxel format describes a 3D object as a matrix of volume occupancy, where the size of the matrix is fixed. Wu et al. \citep{wu2018learning} adopted voxel representation in the generation of 3D shapes. Voxel format requires high resolution to describe fine-grained details; thus, as the shape resolution increases, the computational cost also explodes. The reconstruction results of voxel-based research are limited in resolution and do not provide topological guarantees or represent sharp features.  
As shown in Figure~\ref{fig:rp_b}, point clouds are a lightweight 3D representation composed of $(x, y, z)$ coordinate values. Point clouds are a natural means to represent shapes. PointNet \citep{qi2017pointnet} extracts global shape features using the max-set operations and it is used widely as an encoder for point-based generative networks \citep{voynov2020unsupervised}. However, point clouds do not represent topology and are unsuitable for generating watertight surfaces. Meshes are widely used and constructed from vertices and faces (Figure~\ref{fig:rp_c}). 
\cite{wang2018pixel2mesh} deformed a pre-defined template to restrict a fixed topology using graph convolution. Recently, meshes have been used to represent shapes in deep learning techniques \citep{nash2020polygen}. 
Although meshes are more suitable for describing the topological structure of objects, they usually require advanced preprocessing steps.


\subsubsection{Implicit 3D Shape Representation}

In the field of three-dimensional shape modeling, implicit functions are commonly represented in three ways: occupancy field, signed distance function (SDF), or unsigned distance function (UDF), and the recently emerging neural radiance fields (NeRF). As shown in Figure~\ref{fig:rp_d}, implicit representation refers to describing a surface with a zero-crossing point of a volume function $\psi : R^3 \to R$, whose value can be adjusted.
Representing a 3D shape as a set of level sets of a deep network, mapping 3D coordinates to a signed distance function \citep{Park_2019_CVPR} or occupancy field \citep{Mescheder_2019_CVPR}. Implicit representation can create a lightweight, continuous shape representation with no resolution limits.

Occupancy field is one of the implicit function methods based on deep learning  \citep{Mescheder_2019_CVPR}.
Occupancy field assigns binary values to each point in three-dimensional space, determining whether the point is occupied by an object. This approach utilizes neural networks to learn the representation of occupancy fields, thereby facilitating highly detailed three-dimensional reconstruction. The advantage of occupancy field lies in its dynamic modeling of object occupancy in scenes, thus making it suitable for handling complex three-dimensional environments. Building upon occupancy field, the signed distance function (SDF) has become a crucial direction in implicit function representation within deep learning. SDF assigns a signed distance value to each point, indicating the shortest distance from the point to the object’s surface. Positive values signify points outside the object, while negative values indicate points inside the object.%
As shown in Figure \ref{fig:sdf}. 
DeepSDF \citep{Park_2019_CVPR} provides an end-to-end approach for continuous SDF learning, thereby enabling precise modeling of irregular shapes and local geometry. 

Neural radiance fields (NeRF) \citep{mildenhall2021nerf} have revolutionized the field of computer vision and graphics by introducing a novel approach to scene representation. 
at the heart of NeRF lies the concept of representing a scene as a continuous function capturing radiance information at every point. NeRF introduces an implicit representation, enabling the encoding of detailed and continuous volumetric information. This allows for high-fidelity recon- struction and rendering of scenes with fine-scale structures, surpassing the limitations of explicit representations. Recently, 3D Gaussian Splatting \citep{kerbl3Dgaussians} was introduced by projecting 3D information onto a 2D domain using Gaussian kernels and achieved better performance than NeRF.

\subsubsection{3D Model Generation}
\label{3dg}

\paragraph{Text-to-3D:}
Recent advancements in text-to-3D synthesis have demonstrated remarkable progress, with researchers em- ploying various sophisticated strategies to bridge the gap between natural language descriptions and the creation of detailed 3D content. The pioneering work DreamFusion \citep{poole2023dreamfusion} harnesses a pre-trained 2D text-to-image DM to generate 3D models without large-scale labeled 3D datasets or specialized denoising architectures. Magic3D \citep{Lin_2023_CVPR} improves upon DreamFusion’s \citep{poole2023dreamfusion} limitations by implementing a two-stage coarse-to-fine approach, accelerating the optimization process through a sparse 3D representation before refining it into high-resolution textured meshes via a differentiable renderer. 



\paragraph{Image-to-3D:}
Recent 3D reconstruction techniques specifically focus on generating and reconstructing three-dimensional objects and scenes from a single or few images. NeRF \citep{mildenhall2020nerf} represents a state-of-the-art technique in which complex scene representations are modeled as continuous neural radiance fields optimized with sparse input views. Leveraging the joint language-image embedding space of the CLIP model, CLIP-NeRF \citep{wang2021clip} proposes a unified framework that allows manipulating NeRF in a user-friendly manner, using either a short text prompt or an exemplar image. DreamCraft3D \citep{sun2023dreamcraft3d} introduces a hierarchical process for 3D content creation that employs bootstrapped score distillation sampling from a view-dependent DMs. This two-step method refines textures through personalized DMs trained on augmented scene renderings, thereby delivering high-fidelity, coherent 3D objects. In contrast, Magic123 \citep{Magic123} offers a two-stage solution for generating high-quality textured 3D meshes from unposed wild images. It optimizes a neural radiance field for coarse geometry and fine-tunes details using differentiable mesh representations guided by both 2D and 3D diffusion priors.

\section{Generative AI for Architectural Design Process}

In this paper, it was found that the application of generative AI in architectural design focuses primarily on specific architectural tasks, categorized into concept image generation, architectural 3D form generation, plan generation, facade generation, and structural system generation, as previously mentioned. This paper analyzes and summarizes the application methods for each architectural task, examining input data, output data, application scenarios, and technological approaches.

\subsection{Architectural Concept Design}

\begin{table*}
\centering
\begin{tabular}{ p{3cm} p{3cm} p{3cm} p{6cm} }
\toprule
Input & Output & Application Scenario & Methodology \& Paper \\
\midrule

parameter & concept images & Building & GANs  \citep{kimperspectival}; \\

text & concept images & Building & Text-to-image generative models  \citep{hanafy2023artificial,koehler2023more,paananen2023using,chen2023generating,zhang2023exploringthe,tan2024using,sukkar2024analytical} \\  

text, images & concept images & Building &  DDPM  \citep{chen2023using}; DALLE  \citep{seneviratne2022dalle}; VQGAN,CLIP  \citep{vermisso2022semantic}; StyleGAN  \citep{kim2022latent,chen2021exploration}; GANs  \citep{huang2021gans}; 
Stable-diffusion, Midjourney \citep{zhong2024ai}; VQGAN+CLIP, StyleGAN  \citep{horvath2024ai}; Text-to-image generative model \citep{tong2023can}
\\
\midrule

layout images & urban block layout images & Urban block & GANs  \citep{ye2022masterplangan} \\

interior images & stylistic preference images & Room & GANs  \citep{gui2021gan,shum2023conditional}; \\

sketch images & architectural images & Building & GANs  \citep{qian2022self,xuresearch}; \\

food images & architectural images & Building & GANs,DDPM  \citep{koh2023bewitched}; \\
\midrule

semantic images & concept images & Urban streets & GANs \citep{kimparticipatory}; \\
\midrule
 concept images & semantic images & Landscape & VAE \citep{han2023scan2drawing}; \\
\bottomrule

\end{tabular}
\caption{Application of generative AI in architectural concept image generation.}
\label{tab:image}
\end{table*}

Concept is defined as “the figure of an object, along with other representations, such as attributes or functions of the object, which existed, is existing, or might exist in the human mind, as well as in the real world” \citep{book123}.Concept also refers to the mental representation that the brain uses to denote a class of symbols that are inferred from the physical material \citep{carey2000origin}.
Many architectural concept expressions synthesize design elements into 2D images, thereby reflecting the architect’s personal style and experience.

\paragraph{Application of Generative AI in Architectural Concept Image Generation:}
The applications of generative AI in architectural concept generation include four main categories, as shown in the Table \ref{tab:image}: 
1) generating architectural concept images based on text or images; 2) transfer of architectural concept image style based on images; 3) generating concept images based on semantic images; 4) generating semantic images based on concept images. 

First, \cite{kimperspectival} use linear interpolation techniques to generate architectural images from text across various perspectives, 
while certain authors’ \citep{hanafy2023artificial,koehler2023more,paananen2023using,zhang2023exploringthe,tan2024using,sukkar2024analytical} use of direct generation  from textual prompts simplifies the concept image creation process.
\cite{chen2023generating} exploit the stable DM to generate architectural interior images. Several authors \citep{chen2023using,vermisso2022semantic,kim2022latent,cheng2023concept,seneviratne2022dalle,chen2021exploration,huang2021gans,zhong2024ai,horvath2024ai} use generative AI tools to create architectural design concept images based on image and text prompts.
\cite{tong2023can} incorporated Midjourney into design courses, guiding students to use orthographic projections as input to generate conceptual images.

Second, 
\citep{ye2022masterplangan} used a GAN model to colorize line sketches of urban block layout images.
\cite{gui2021gan} exploit GAN models to generate comfortable underground space renderings from virtual 3D space images; \cite{shum2023conditional} utilize GANs to facilitate the creation of interior decoration images from 360-degree panoramic interior images. 
\cite{qian2022self} and \cite{xuresearch} utilize GANs to generate architectural images based on sketch line drawings.
\cite{koh2023bewitched} explores the use of GANs and DDPM for style transfer on food images, converting them into creative architectural images.

Third,\cite{kimparticipatory} use GANs to generate urban street images based on image semantic labels (masks), allowing precise control over the content of the generated images. \cite{han2023scan2drawing} use VAE to produce semantic images corresponding to architectural images.

\subsection{Architectural 3D Forms Design}

According to \cite{ching2023architecture} , “in art and design, we often use the ‘form’ to denote the formal structure of a work the manner of arranging and coordinating the elements and parts of a composition so as to produce a coherent image.” Architectural 3D form design refers to the initial stage of designing 3D conceptual models or representations of buildings or structures. This phase focuses on exploring and developing basic spatial configurations, volumes, and massing of the architectural design before detailed refinement.

\begin{table*}
\centering
\begin{tabular}{ p{3cm} p{3cm} p{3cm} p{6cm} }
\toprule
Input & Output & Application Scenario & Methodology \& Paper \\
\midrule

parameters & architectural 3D forms & Building volumes & VAE \citep{de2020generation}; GAN, VAE \citep{zhuang2023synthesis};
3D-DDPM \citep{liu2023diffusion}; DNN \citep{zheng2021generative}; 
3D-GAN, CPCGAN \citep{pouliou2023speculative}; CVAE \citep{sebestyen2023using}; DCGAN, StyleGAN \citep{mayrhoferlinear}; 3D-GAN \citep{ennemoser2023design}; 3D-GAN \citep{CAADRIA2023} \\

architectural 3D forms & classify & Building volumes & 3D-AAE \citep{kimtowards}; VAE \citep{hasey2023form} \\
text & architectural 3D forms & Urban block volumes & LDM \citep{zhuang2024text} \\
\midrule

sketches & architectural 3D forms & Building volumes & VAE, GAN \citep{tono2022vitruvio} \\
Color-Patterns & architectural 3D forms & Building volumes & CGAN \citep{navarro2021color}\\ 
sketch, environmental images & architectural 3D forms & Urban block volumes & GAN \citep{kim2020citycraft} \\
volumes & architectural 3D forms & Building volumes & DDPM \citep{sebestyen2023generating} \\
floor plan & architectural 3D forms & Building volumes & StyleGAN \citep{asmar2020machinic} \\
spatial sequence diagrams, spatial requirements & architectural 3D forms & Building volumes & CGAN, GNN \citep{inproceedings} \\
\midrule

images with 3D form information & images with 3D form information & Building volumes & pix2pix \citep{zhang2021machine}\\ 

site conditions, surroundings conditions & images with 3D form information & Urban block volumes &
DCGAN \citep{quan2022urban}; pix2pix, CycleGAN \citep{zhou2023automatic,li2023study,vesely2022building,jiang2023building}\\
\midrule

architectural 3D forms, site conditions & heatmap of environmental performance evaluation & Building volumes performance evaluation & VAE \citep{ampanavos2021early}; pix2pix, cycleGAN \citep{huang2022accelerated} \\

\bottomrule
\end{tabular}
\caption{Applications of generative AI in architectural 3D forms generation.}
\label{tab:P-3D}
\end{table*}

\paragraph{Applications of Generative AI in Architectural 3D Forms Generation:}

The applications of generative AI in this process include four main categories, as shown in the Table \ref{tab:P-3D}: 
1) generating architectural 3D forms based on parameters or text, and classification of preliminary 3D forms; 2) generating architectural 3D forms based on images (usually from sketches, color patterns, floor plans, etc.) and voxel volume; 3) generating images with 3D form information; 4) generating environmental performance evaluation based on 3D forms.

First, generative AI facilitates the generation of preliminary 3D forms based on input parameters or the conduct of classification analysis. Initially, \cite{de2020generation} use VAEs for generating preliminary 3D forms from input parameters. Building on this, \cite{zheng2021generative} apply GANs to refine the process by training point cloud data of 3D models and using category prompts for more precise reconstructions. 
\cite{zhuang2023synthesis} employ the approach that facilitates the creation of innovative architectural 3D forms using input interpolation. 
\cite{liu2023diffusion} utilize diffusion probability models to provide a unique training method for Taihu stone and architectural 3D forms and discover transitional forms through interpolation. 
\cite{pouliou2023speculative} use a structural GAN model that uses point cloud data to generate 3D models based on parameters such as length, width, and height. 
\cite{sebestyen2023using} exploit VAEs to generate 3D voxel models guided by textual labels. 
\cite{mayrhoferlinear}  apply interpolation to create new 3D elements with transitioning forms.
\cite{ennemoser2023design} operate 2D images with 3D voxel information, which can be produced using the input RGB channel values.
\cite{CAADRIA2023} use 3D-GAN to generate voxelized and point cloud representations of building 3D forms components in accordance with textual labels.
\cite{kimtowards} employ the 3D adversarial autoencoder model to train point cloud representations of a 3D model, thereby facilitating the classification of architectural forms.
\cite{hasey2023form} utilize VAE to train signed distance function (SDF) voxels and to conduct clustering analysis on latent vector representations of 3D models.
\cite{zhuang2024text} present a novel method to generate urban block depth images from textual descriptions using latent DMs, particularly stable diffusion, and reconstruct 3D urban models from these depth images.

Second, generative AI involves using images as the generation conditions for generative AI to produce 3D forms.
\cite{tono2022vitruvio} utilize the integration of VAE and GAN models to facilitate the generation of architectural 3D forms from sketches.
\cite{navarro2021color} operate CGAN to create architectural 3D models from design concept sketches, and \cite{kim2020citycraft} generate 3D models from a singular concept sketch combined with environmental images.
\cite{sebestyen2023generating} utilize diffusion probability models to train 3D models, introducing noise into the 3D voxel volume to create novel forms.
\cite{asmar2020machinic} employ StyleGAN to generate detailed 3D forms from architectural floor plans. In addition,
\cite{inproceedings} use CGAN to generate detailed 3D forms based on spatial sequence diagrams and spatial requirements.

Third, training on 3D model data is more challenging than that on 2D image data. Researchers have simplified the training process to address these challenges by converting 3D forms into 2D image representations, such as grayscale images enriched with height information. 
\citep{zhang2021machine} exploit the practice of transforming 3D models into section images for reconstruction, followed by reverting these section images back to 3D forms, thereby significantly reducing both training duration and costs and ensuring accurate restoration of the original 3D models. 
A few authors \citep{quan2022urban, zhou2023automatic,li2023study,vesely2022building,jiang2023building} use this method for the generation of architectural 3D forms tailored to specific sites and their surrounding environments.

Fourth, \cite{ampanavos2021early} and \cite{huang2022accelerated} exploit generative AI to conduct site and architectural environmental performance evaluations based on 3D models. This involves generating images for assessments, such as view analysis, sunlight exposure, and solar radiation, among others.

\subsection{Architectural Plan Design}

An architectural plan is a horizontal section taken at approximately different levels \citep{wakita2003professional}. Architectural plan design includes building floor plan design and layout design. Building floor plan design refers to the layout or plan view of a specific structure on a horizontal plane. It depicts the detailed arrangement of different levels of the building (typically ground floor, upper floors, etc.), including the positions, dimensions, and relationships of rooms, corridors, walls, doors, and windows. Further, architectural floor plan design emphasizes spatial functional zoning, scale proportions, room uses, and other aspects , making it a fundamental and crucial component of architectural design. Layout design typically refers to the broader arrangement and configuration of spaces or sites, not limited to the interior of buildings. It can encompass the positioning of buildings on a site, relationships among buildings, and the arrangement of other elements such as roads and landscapes within the overall environment. Layout design focuses on optimizing space utilization, enhancing functional efficiency, and meeting design requirements and environmental conditions while considering the interaction between buildings and their surroundings.


\begin{table*}
\centering
\begin{tabular}{ p{3cm} p{3cm} p{3cm} p{6cm} }
\toprule
Input & Output & Application Scenario & Methodology \& Paper \\
\midrule

building footprint,functional space layout & functional space layout,floor plan & Building & GANs \citep{chaillou2020archigan,gao2023m,min2023floor,wan2023deep,chen2023generative};\\

microorganism drawings & floor plan & Building &  CycleGAN \citep{akdougan2022plan} \\

floor plan & flat furniture layouts & Room & pix2pix \citep{karadag2023ai}; CGAN \cite{tanasra2023automation} \\

floor plan & floor plan & Building & DCGAN \citep{uzun2020gan} \\

environmental performance evaluation & floor plan & Building & pix2pix \citep{huang2023damascus} \\

historical maps & satellite imagery & Urban block & CycleGAN \citep{alaccam2022reciprocal} \\

\midrule

functional space layout & functional space layout & Urban block & GNN, CVAE \citep{xu2021blockplanner}; GANs \citep{yang2023street} \\

functional space layout, requirements and standards & functional space layout & Building & Transformer \citep{hosseini2022floorplan}; CoGAN \citep{ghannad2022automated}; GC-GAN \citep{liu2024intelligent}  \\

spatial sequences & functional space layout & Building & GANs \citep{nauata2020house,nauata2021house,wang2021actfloor,velosopedagogical,luo2022floorplangan,aalaei2023architectural}; DDPM \citep{shabani2023housediffusion} \\

site condition & functional space layout & Campus & CGAN \citep{liu2022exploration,liu2021exploration} \\

site condition & functional space layout & Urban block & pix2pix \citep{sun2023development} \\

building footprints & functional space layout & Building & CGAN \citep{zhao2021generation}; StyleGAN, Graph2Plan, RPLAN, HouseGAN \citep{para2021generative} \\

bubble chart & functional space layout & Building & CGAN \citep{han2024graph2pix} \\
bubble charts, building footprints, designer requirements & functional space layout & Building & DDPM \citep{zeng2024residential} \\
designer requirements & functional space layout & Building & LLMs \citep{lichatdesign} \\
satellite imagery & urban functional layout & Urban block & DCGAN, Style-GAN \citep{mayrhofer2023advancing} \\
\midrule

floor plan & spatial sequences & Building & EdgeGAN \citep{dong2021vectorization} \\

isovists & wall Spatial sequences & Building & VQ-VAE, GPT \citep{johanes2023generative} \\

site conditions & spatial sequences & Building & DDPM \citep{su2023floor} \\

\midrule

floor plan & heatmaps of space environmental performance evaluation & Building & CGAN \citep{Doumpioti2022FieldC} \\

functional space layout & heatmaps of space environmental performance evaluation & Building & pix2pix \citep{mostafavi2022interactive}; pix2pix \citep{he2021predictive} \\
\bottomrule

\end{tabular}
\caption{Application of generative AI in the architectural plan generation.}
\label{tab:Plan}
\end{table*}

\paragraph{Application of Generative AI in Architectural Plan Generation:}
The applications of generative AI in architectural plan generation include four main categories, as shown in the Table \ref{tab:Plan}: 
1) generating building floor plans based on 2D images, usually from building footprints, heatmap of environmental performance evaluation, functional space layout, spatial sequences, and floor plan; 2) generating functional space layout based on 2D images , usually from functional space layout, spatial sequences, building footprints, requirements and standards, site conditions, and surroundings conditions; 3) generating spatial sequences based on 2D images, usually from building floor plan, building footprints, and environmental performance evaluation heatmap; 4) generating spatial environmental performance evaluations heatmap based on 2D images, usually from building floor plans and functional space layout.

First, in generating architectural floor plans, certain authors \citep{gao2023m,min2023floor,chen2023generative,wan2023deep,chaillou2020archigan} create a functional space layout diagram from building footprints or site boundaries and then develop an architectural floor plan, progressing from building footprint or site boundaries to functional space layout, and finally to floor plan. 
\cite{akdougan2022plan} use generative AI models to convert microorganism drawings into architectural floor plans in Palladio’s style. 
\cite{karadag2023ai} and \cite{tanasra2023automation} utilize GANs model to refine architectural floor plans, obtaining plans with flat furniture layouts. 
The generation and reconstruction of floor plans are achieved by  \cite{uzun2020gan} through training on architectural floor plan datasets.
\cite{huang2023damascus} generate floor plans based on the heatmap of spatial environmental evaluation, such as wind and lighting conditions.
\cite{alaccam2022reciprocal} employed CycleGAN to perform bidirectional conversions between Istanbul's historical Pervititch maps and plan views derived from modern satellite imagery, facilitating plan generation based on the transformed views.

Second, generative AI plays a various roles in the generation of functional space layouts. 
\cite{xu2021blockplanner} exploit generative AI to reconstruct and produce matching functional layout diagrams based on the implicit information within the functional space layout.
\cite{yang2023street} generate street layout based on functional space layout. Certain researchers\citep{hosseini2022floorplan,ghannad2022automated,liu2024intelligent} generate or complete incomplete functional space layouts based on specific demands. 
Many researchers \citep{nauata2020house,nauata2021house,wang2021actfloor,velosopedagogical,luo2022floorplangan,shabani2023housediffusion,aalaei2023architectural} use spatial sequence diagrams to generate functional space layout. 
Moreover, several authors \citep{sun2023development,liu2022exploration,liu2021exploration} use generative AI to generate functional space layouts based on the designated site boundary.
\cite{zhao2021generation} and \cite{para2021generative} skillfully use building footprints as conditions to generate a functional space layout. 
\cite{han2024graph2pix} proposes a graph-based deep learning model called graph2pix (G2P), which incorporates room area and type information into the graph’s nodes (bubble chart) to generate floor plans that meet specific user requirements.
\cite{zeng2024residential}, based on DDPM, introduce a multi-conditional generative model called FloorplanDiffusion; by inputting bubble charts, building footprints, and designer requirements, the model generates diverse, high-quality, and controllable residential floor plans.
\cite{lichatdesign} introduces ChatDesign, a method utilizing pre-trained LLMs to generate functional space layouts from natural language descriptions. 
\cite{mayrhofer2023advancing} use DCGAN and StyleGAN to convert satellite image datasets into urban functional layout images.

Third, generative AI models demonstrate exceptional performance in the generation of spatial sequences. \cite{dong2021vectorization} utilize EdgeGAN to identify and reconstruct wall layout sequences from floor plans.  
\cite{johanes2023generative} employ isovists to predict wall sequence diagrams.
\cite{su2023floor} use a DDPM to generate spatial sequence diagrams based on specific boundaries. 
Fourth, \cite{Doumpioti2022FieldC} utilize CGAN to foresee space environmental performance evaluations from floor plans, such as light exposure and isovists. \cite{mostafavi2022interactive} utilize pix2pix to predict indoor brightness  
and \cite{he2021predictive} utilize pix2pix to predict solar radiation based on functional space layout diagrams.

\subsection{Architectural Facade Design}

Architectural facade design refers to the process of designing the exterior appearance of a building, specifically focusing on the facade or the outer shell. The facade plays a crucial role in the overall aesthetic appeal of a building and often serves functional purposes such as providing weather protection, insulation, and visual identity. Facade design involves considerations of materials, textures, colors, proportions, and architectural styles to achieve the desired visual impact while meeting functional requirements \citep{herzog2004facade}.

\begin{table*}
\centering
\begin{tabular}{ p{3cm} p{3cm} p{3cm} p{6cm} }
\toprule
Input & Output & Application Scenario & Methodology \& Paper \\
\midrule

semantic segmentation maps of facade & facade images & Building & GANs \citep{du20203d,yu2020architectural,chuang2021facilitating,sun2022automatic,zhang2022cgan,lin2023research,zhangsynthesizing,jo2024generative}\\
\midrule

facade images & facade images & facade style transfer & CycleGAN \citep{dongurban}; StyleGAN2 \citep{meng2022exploring}; GANs \citep{cciccek2023deterioration}; Neural style transfer \cite{sun2022application} \\
\midrule

architectural facade outline images & semantic segmentation maps of facade & Building & pix2pix \citep{wan2023deep} \\

incomplete semantic segmentation maps of facade & semantic segmentation maps of facade & Building & GAN \citep{cai2021building} \\

\bottomrule
\end{tabular}
\caption{Application of generative AI in architectural facade generation.}
\label{tab:Facade}
\end{table*}

\paragraph{Application of Generative AI in Architectural Facade Generation:}
The applications of generative AI in architectural facade design include three main categories, as shown in the Table \ref{tab:Facade}: 
1) generating facade images based on semantic segmentation maps; 2) using generative AI models for facade image style transfer; 3) generating semantic segmentation maps of facade based on images. 

In generating facade images, many researchers \citep{yu2020architectural,chuang2021facilitating,sun2022automatic,zhang2022cgan,lin2023research,zhangsynthesizing,du20203d,jo2024generative} utilize generative AI to create architectural facade images based on semantic segmentation maps that precisely annotate the form and location of elements such as walls, windows, and other components. 
In facade style transfer, \cite{dongurban} use CycleGAN to train facade datasets for generating novel architectural facade images.  \cite{sun2022application} and \cite{cciccek2023deterioration} apply style transfer to architectural facades by incorporating style images. \cite{meng2022exploring} facilitates style transfer between facade images to generate a new facade image.
In generating semantic segmentation maps of architectural facades, \cite{cai2021building} utilize GANs to reconstruct these maps, including rebuilding occluded parts from unobstructed areas. 
\cite{wan2023deep} utilize pix2pix to complete facade mask images for all four building directions, which can be generated from images of the outline of the architectural facade.

\subsection{Architectural Structural System Design}

Architectural structural system design refers to the process of designing the structural framework and system that sup- ports a building’s architecture. It involves determining the type of structural elements (such as beams, columns, slabs, and walls), their arrangement, and their integration with the architectural design to ensure structural stability, safety, and functionality of the building \citep{allen2019fundamentals}.

\begin{table*}
\centering
\begin{tabular}{ p{3cm} p{3cm} p{3cm} p{6cm} }
\toprule
Input & Output & Application Scenario & Methodology \& Paper \\
\midrule

floor plan & structural layout & Building & GANs \citep{liao2021automated,fei2022integrated,fu2023dual,kosenciug2024structural} \\

floor plan, structural load capacities & structural layout & Building & GANs \citep{liao2022intelligent,lu2022intelligent,fei2023semi,zhao2023intelligent} \\

structural layout & structural layout & Building (Structural optimization) & GANs \citep{liao2023base} \\

functional space layout, floor plan & structure layout & Building & pix2pixHD \citep{zhao2022intelligent} \\

floor plan & structure layout & Building & DDPM \citep{gu2024intelligent} \\
\midrule

structural layout, structural dimensions & structural layout, structural dimensions & Building (Structural optimization) & StructGAN-KNWL \citep{fei2022knowledge} \\
 
\bottomrule
\end{tabular}
\caption{Application of generative AI in architectural structural system generation.}
\label{tab:str}
\end{table*}

\paragraph{Application of Generative AI in Architectural Structural System Generation:}
The applications of generative AI in structural system design primarily involve the prediction and optimization of structural layout and dimensions. 
The application methods are illustrated in Table \ref{tab:str}.
In predicting structure layout, several researchers \citep{liao2021automated,fei2022integrated,fu2023dual,kosenciug2024structural} utilize generative AI to recognize architectural floor plans and generate structural layout images.
A few authors \citep{lu2022intelligent,liao2022intelligent,fei2023semi,zhao2023intelligent} exploit generative AI to generate structural layout diagrams based on specified structural load capacities. 
Generative AI can refine existing structural layouts. \cite{liao2023base} use GANs to propose seismic isolation solutions for walls, and \cite{zhao2022intelligent} use pix2pixHD to combine functional space layout and floor plan to create the corresponding structure layout.  
\cite{gu2024intelligent} propose an intelligent shear wall layout design method based on DMs.
In optimizing structural layout, \cite{fei2022knowledge} apply StructGAN-KNWL to forecast and create more suitable structural sizes and layouts based on existing structural layouts.

\subsection{Architectural Section Design}
The building section is the premier drawing for revealing and studying the relationship between the floors, walls, and roof structure of a building and the dimensions and vertical scale of the spaces defined by these elements \citep{ching2023architectural}.

\paragraph{Application of Generative AI in Architectural Section Generation:}
The applications of generative AI in architectural section design include three main categories, as shown in the Table \ref{tab:Section}: 
1) spatial feature extraction and construction from section; 2) style transfer and form generation from section; 3) architectural restoration and reconstruction from section. 
First, \cite{deng2023exploration} use Pix2Pix to generate sequentially stacked section drawings of Taihu stone, extract spatial variation patterns, and construct a 3D form reflecting Taihu stone characteristics. Second, \cite{zhang20203d} decompose 3D architectural models into 2D sequentially stacked section images and applied pix2pix and cycleGAN, to modify the style of the section drawings.They generated new stylized section images and used them to construct 3D architectural forms, ensuring pixel continuity to preserve the model's structural and spatial integrity. Third, \cite{guzelci2022machine} use pix2pix to train on collected kümbet section drawings and employed SSIM (Structural Similarity Index) to evaluate image similarity, facilitating the prediction and reconstruction of missing architectural components.

\begin{table*}
\centering
\begin{tabular}{ p{3cm} p{3cm} p{3cm} p{6cm} }
\toprule
Input & Output & Application Scenario & Methodology \& Paper \\
\midrule

section image & section image & Building volume & pix2pix \citep{deng2023exploration} \\
\midrule
section image & section image & Building volume style transfer & pix2pix, cycleGAN \citep{zhang20203d} \\
\midrule
section image & section image & Building volume restoration & pix2pix \citep{guzelci2022machine} \\

\bottomrule
\end{tabular}
\caption{Application of generative AI in architectural section generation.}
\label{tab:Section}
\end{table*}

\section{Future Research Directions}

In this section, we summarize the current generative AI models of image, video, and 3D forms that could be applied to six architectural design steps in the future.
According to the research discussed in this paper (Section~\ref{sec:gam}), generative AI technology can be categorized into image generation models, video generation models, 3D generation models, and language generation models. This classification is derived from the outputs of various generative AI models. 
Since language generation models are not directly applicable in architectural design, the focus of current and future application technologies lies in image, video, and 3D generation models. Publicly available data from the Huggingface platform indicate that the number of video and 3D generation models is considerably smaller than that of image generation models. 
Image generation models are predominantly used by architects due to their advanced technology. As technology progresses, their application in architectural design is expected to be more effective and widespread in the future. 

Currently, generative AI is limited to specific design steps and cannot yet handle the full architectural design process. In the future, technological advancements could lead to a comprehensive generative AI platform, allowing designers to flexibly select and combine modules, resulting in a more adaptable and integrated design process.
Data consistency and compatibility are essential to ensure seamless information transfer between different steps in architectural generative AI platform. Leveraging the cross-step capabilities of generative AI platform can effectively link design steps by using the output of one step as the input for the next. 
For example, the conceptual design step can generate concept images based on the designer's ideas by text/image-to-image models, and text/image-to-3D model generation technology can directly transform design conceptual images into initial 3D models, providing a foundation for detailed design. Floor plans and elevations designs can be automatically extracted from these 3D models. Then, AI models can generate detailed drawings that meet design specifications. In the structural and sectional design step, structural analysis and sectional drawings can also be automatically performed based on the generated 3D model, ensuring the design's feasibility and safety.
A real-time synchronization mechanism ensures that changes in one design step are instantly reflected in others, maintaining consistency and efficiency.

\subsection{Future Research Directions for Architectural Concept Generation}

Presently, text-to-image generation models produce creative architectural Concept images from brief descriptions or specific parameters, while image-to-image models generate images with consistent styles or features. These models enable the exploration of architectural forms beyond current human conceptions. 
Text/Image-to-image models such as DALL·E 3 \citep{ramesh2022hierarchical}, ControlNet \citep{zhang2023adding}, and SceneGenie \citep{farshad2023scenegenie} offer style transfer and scene control but still struggle to blend complex styles. Future improvements should focus on better transferring architectural styles, materials, and forms, enabling more flexible exploration of design concepts. Enhancing AI's multimodal input and integrating natural language descriptions, 2D sketches, and 3D models will streamline conceptual design generation.

\begin{figure*}[htbp]
    \centering
    \includegraphics[width=0.95\linewidth]{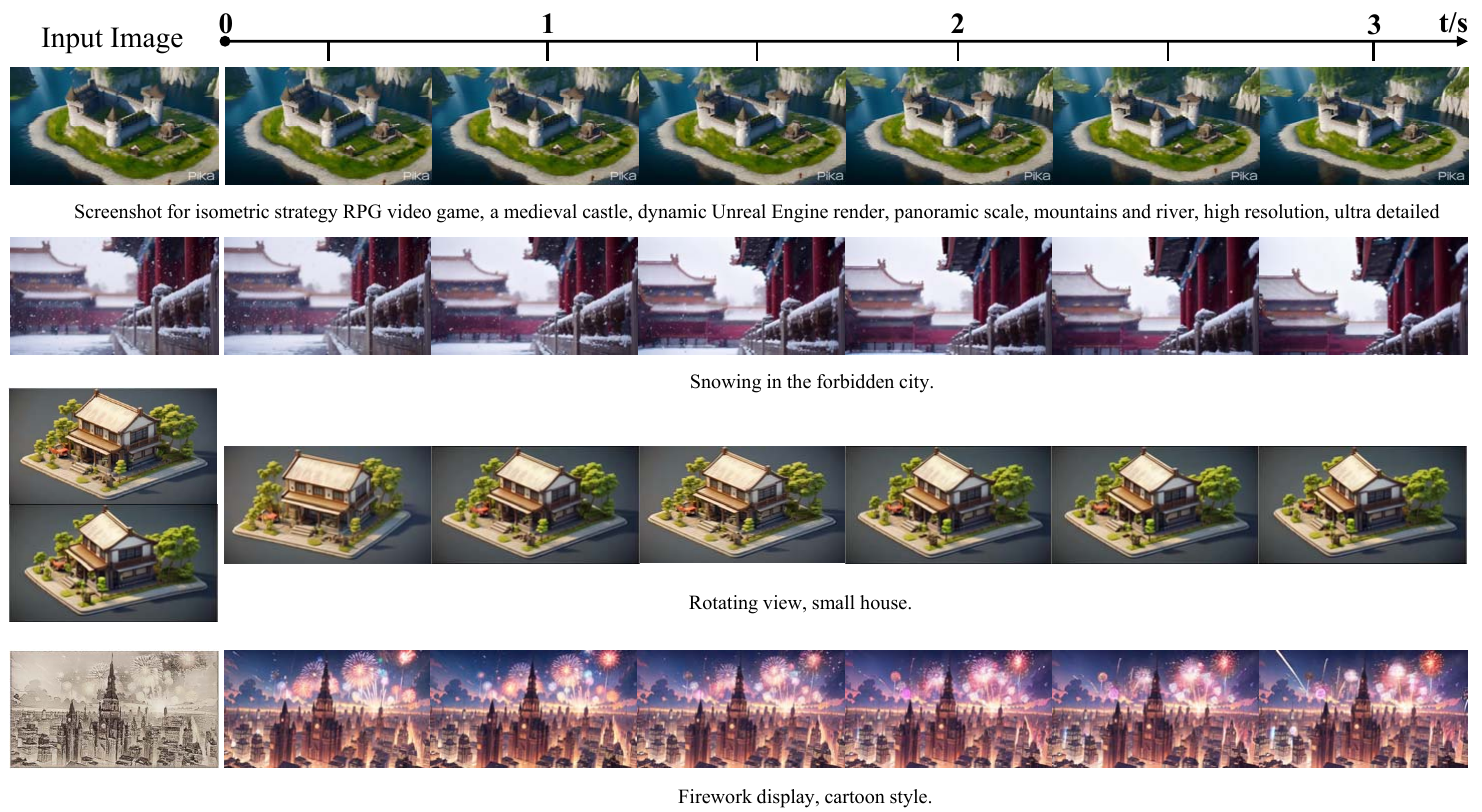}
    \caption{Currently, generative models, such as PIKA \citep{wang2023pika} and DynamiCrafter \citep{xing2023dynamicrafter}, are capable of generating high-quality videos from images, supporting multi-angle rotation, and style transfer.}
    \label{fig:4.2}
\end{figure*}

Generative AI-based video models open new possibilities for architectural concept generation. These models create dynamic visualizations from a single architectural image or textual description, enabling architects to refine concepts in motion.
As shown in Figure~\ref{fig:4.2}, the first and second rows depict effect demonstration videos generated from input images using PIKA \citep{wang2023pika}, where buildings undergo minor movements and scaling while maintaining consistency with the surrounding environment. DynamiCrafter \citep{xing2023dynamicrafter} can generate rotating buildings, as demonstrated in the third row, where the model predicts architectural styles from different angles and ensures consistent generation. The application of these video models expands the possibilities for architectural concept generation by allowing designers to explore and present ideas dynamically.

Maintaining fine details across frames is crucial in architectural concept videos, especially for materials, textures, and structural elements. Make-A-Video \citep{singer2022make} converts text descriptions into dynamic videos, but its ability to preserve detail and ensure smooth transitions between frames needs improvement. In the future, refining temporal convolution and attention mechanisms could enhance frame-by-frame consistency, ensuring that intricate architectural elements like textures and lighting transition throughout the video sequences.
DynamiCrafter \citep{xing2023dynamicrafter} adds dynamic effects like moving clouds and flowing water to static architectural images. Future advancements could incorporate physics-based algorithms to simulate realistic environmental interactions, such as wind affecting facades or light shifting throughout the day. This would significantly enhance the realism and functionality of architectural concept videos, allowing architects to test and visualize how their designs interact with the environment.
Although DynamiCrafter supports high-resolution models, further enhancements could bring video quality closer to photorealism, which is crucial for presenting architectural concepts. Incorporating dynamic elements such as human activity, vehicles, and environmental changes would enrich the architectural narrative, providing a more immersive experience.
PIKA \citep{wang2023pika} adjusts elements in architectural videos based on prompts while maintaining overall image integrity. In the future, it could improve in handling complex scene transformations. Architectural design often requires structural changes or material replacements, so enhancing PIKA to manage large-scale modifications would be beneficial. Using advanced prompt-to-scene transformation techniques, PIKA could allow significant design changes without compromising building integrity, leading to smoother and more natural presentations of architectural designs in video form.

\begin{figure}[t]
    \centering
    \begin{subfigure}{0.15\linewidth}
        \includegraphics[width=\linewidth, height=3cm]{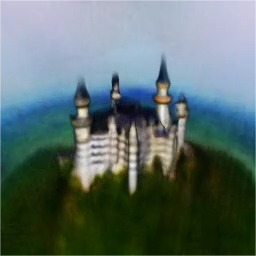}
        \caption{Castle}
        \label{fig:dreamfusion_a}
    \end{subfigure}
    \hspace{0.01\linewidth} 
    \begin{subfigure}{0.15\linewidth}
        \includegraphics[width=\linewidth, height=3cm]{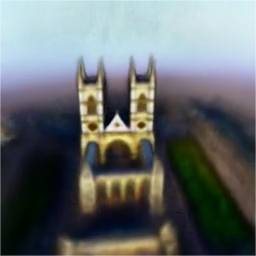}
        \caption{Abbey}
        \label{fig:dreamfusion_b}
    \end{subfigure}
    \hspace{0.01\linewidth}
    \begin{subfigure}{0.15\linewidth}
        \includegraphics[width=\linewidth, height=3cm]{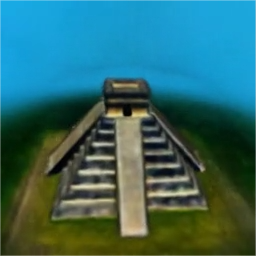}
        \caption{Chichen Itza}
        \label{fig:dreamfusion_c}
    \end{subfigure}
    \hspace{0.01\linewidth}
    \begin{subfigure}{0.15\linewidth}
        \includegraphics[width=\linewidth, height=3cm]{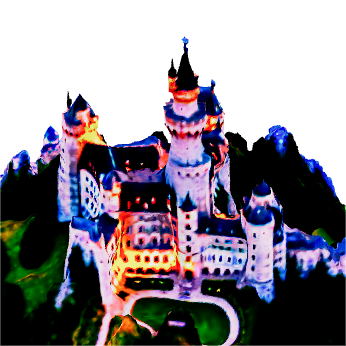}
        \caption{Castle}
        \label{fig:magic3d_a}
    \end{subfigure}
    \hspace{0.01\linewidth}
    \begin{subfigure}{0.15\linewidth}
        \includegraphics[width=\linewidth, height=3cm]{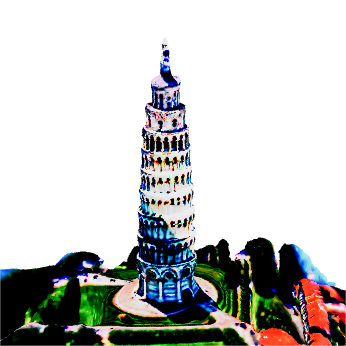}
        \caption{Pisa tower}
        \label{fig:magic3d_b}
    \end{subfigure}
    \hspace{0.01\linewidth}
    \begin{subfigure}{0.15\linewidth}
        \includegraphics[width=\linewidth, height=3cm]{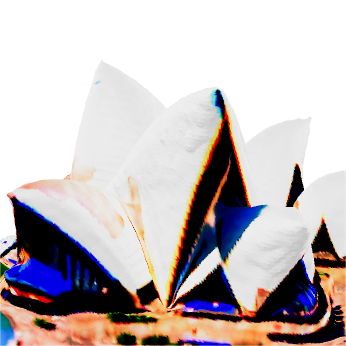}
        \caption{Opera house}
        \label{fig:magic3d_c}
    \end{subfigure}
    \caption{Examples of 3D models generated based on textual prompts. Figures (a) -- (c) are buildings produced by Dreamfusion \citep{poole2023dreamfusion}, figures (d) -- (f) are buildings generated using Magic3D \citep{Lin_2023_CVPR}.}
    \label{fig:ex_4.3_1}
\end{figure}

\subsection{Future Research Directions for Architectural 3D Forms Generation}

\begin{figure}[t]
    \centering
    \begin{subfigure}{0.45\linewidth}
        \centering
        \includegraphics[width=\linewidth, height=4cm]{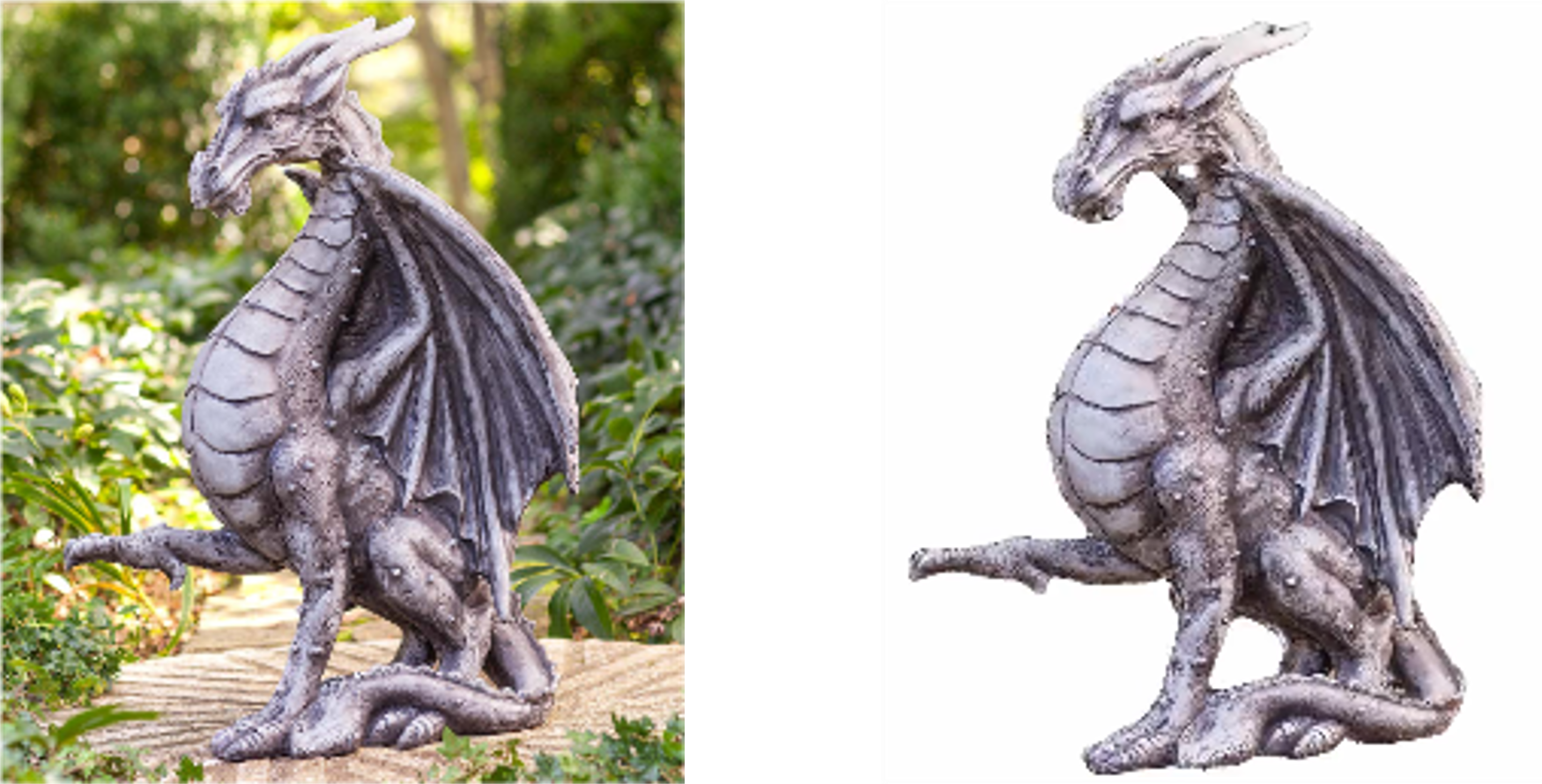}
        \caption{Dragon: input image (left) and 3D model (right)}
        \label{fig:magic123_a}
    \end{subfigure}
    \hspace{0.05\linewidth} 
    \begin{subfigure}{0.45\linewidth}
        \centering
        \includegraphics[width=\linewidth, height=4cm]{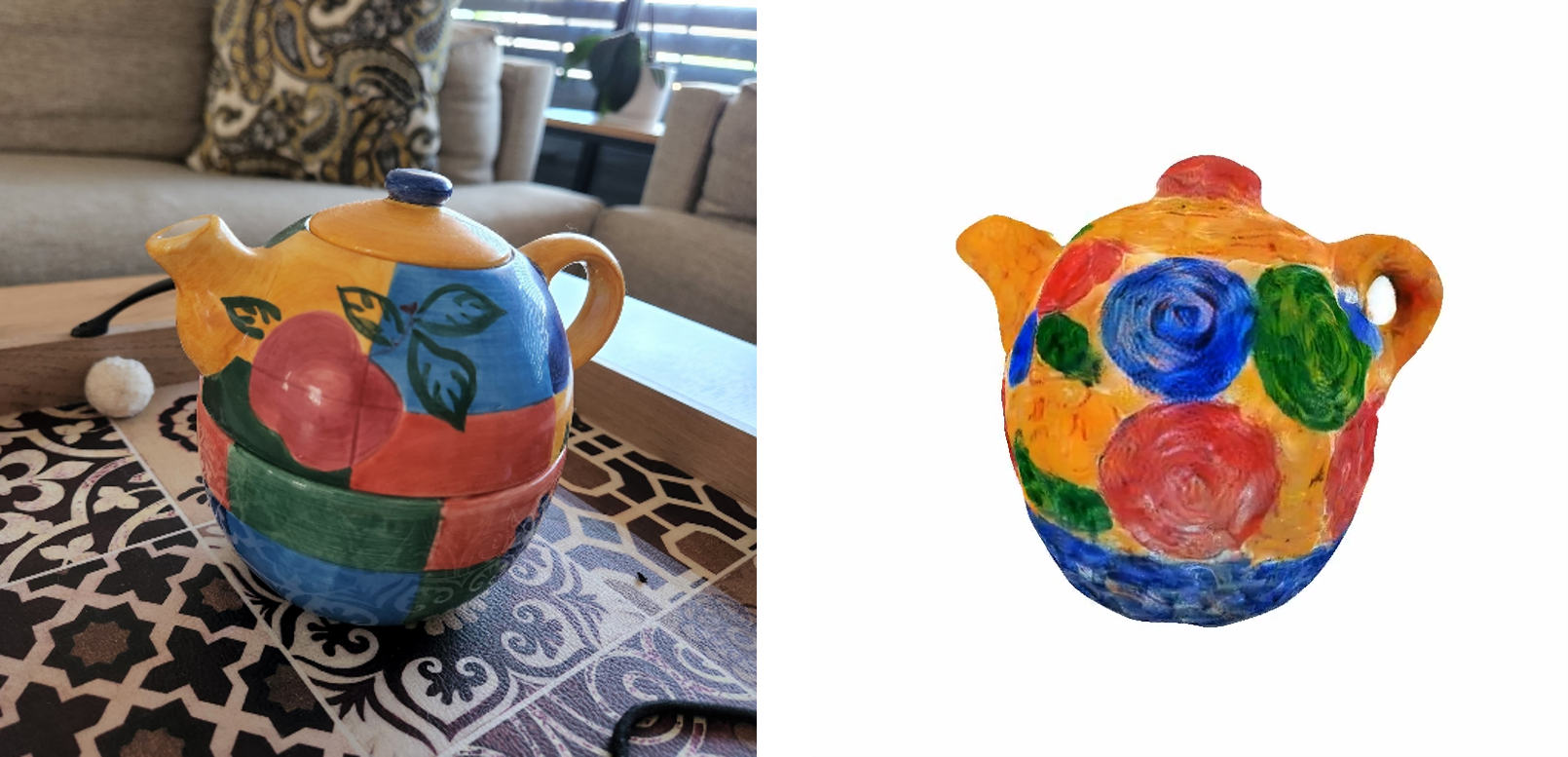}
        \caption{Teapots: input image (left) and 3D model (right)}
        \label{fig:magic123_c}
    \end{subfigure}
    \caption{Examples of input images and the corresponding detailed 3D models generated using Magic123 \citep{Magic123}.}
    \label{fig:ex_4.3_3}
\end{figure}

\begin{figure}[t]
    \centering
    \begin{subfigure}{0.45\linewidth}
        \includegraphics[width=\linewidth, height=3cm]{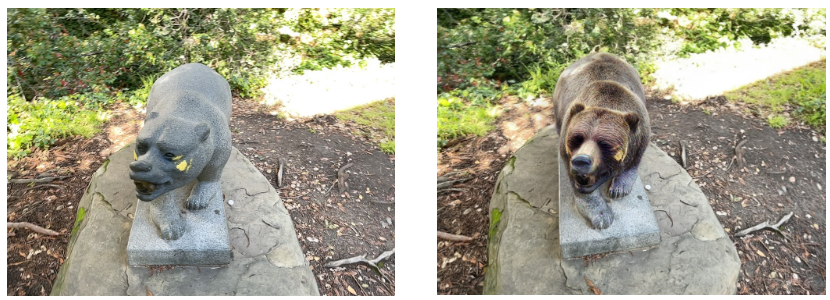}
        \caption{Turn the bear into a Grizzly bear}
        \label{fig:gaussianeditor_a}
    \end{subfigure}
    \hspace{0.05\linewidth} 
    \begin{subfigure}{0.45\linewidth}
        \includegraphics[width=\linewidth, height=3cm]{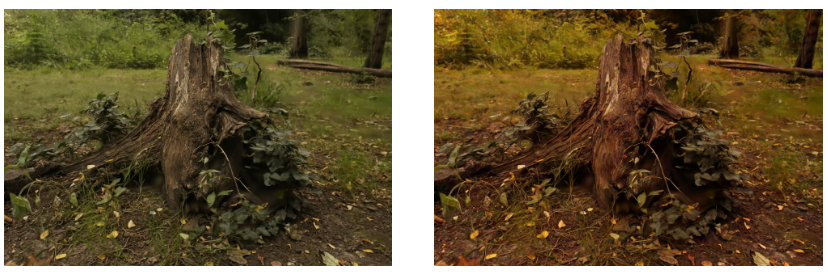}
        \caption{Make it Autumn}
        \label{fig:gaussianeditor_b}
    \end{subfigure}
    \caption{Examples of 3D model style editing using GaussianEditor \citep{chen2023gaussianeditor}. Left column shows the original models, right column presents the edited results.}
    \label{fig:ex_4.3_2}
\end{figure}

Nowadays, using architectural images or text prompts as inputs to generate 3D architectural forms can enhance modeling efficiency. 
As shown in Figure~\ref{fig:ex_4.3_1}, DreamFusion \citep{poole2023dreamfusion} and Magic3D \citep{Lin_2023_CVPR} facilitate the rapid creation of 3D architectural models from text descriptions.
Magic123 employs a two-stage process that transforms complex real-world images into detailed 3D models 
, illustrated in Figure~\ref{fig:ex_4.3_3}. 
GaussianEditor \citep{chen2023gaussianeditor} excel in generating and refining architectural 3D models. 
GaussianEditor utilizes Gaussian semantic tracing and Hierarchical Gaussian Splatting for precise and intuitive architectural detail editing, as depicted in Figure~\ref{fig:ex_4.3_2}. 

DreamFusion \citep{poole2023dreamfusion} and Magic3D \citep{Lin_2023_CVPR} generate 3D architectural models quickly but need improvement in high-resolution and detailed rendering. Current models need more clarity in handling complex details, textures, and fine structures. In the future, enhancing rendering capabilities will improve detail preservation, resulting in richer visual presentations, which is essential for accurately conveying architectural designs.
CLIP-NeRF \citep{wang2021clip} and GaussianEditor \citep{chen2023gaussianeditor} offer editing features that allow architects to adjust models using text or image prompts. In the future, real-time interaction and editing can be improved. Future technologies could enable architects to adjust specific model details in real-time, such as dynamically modifying facades or window layouts through a more intuitive interface. This would significantly accelerate the design process and improve efficiency.
DreamCraft3D \citep{sun2023dreamcraft3d} generates personalized 3D models from text and image prompts, but personalization remains limited. Future models could enhance customization by allowing designers to input detailed parameters like room layouts, materials, and styles, generating diverse 3D forms tailored to specific needs. Supporting more complex prompts would enable a more flexible exploration of architectural forms and styles.
Current models rely on single-mode inputs like text or images, but future models could support diverse inputs such as architectural floor plans and facade details for more precise 3D models. Integrating multi-modal inputs would better capture creativity while considering practical construction needs, enabling architects to make more comprehensive plans in the early design stages.


\subsection{Future Research Directions for Architectural Floor Plan Generation}

\begin{figure}[htbp]
    \centering
    \includegraphics[width=0.6\linewidth]{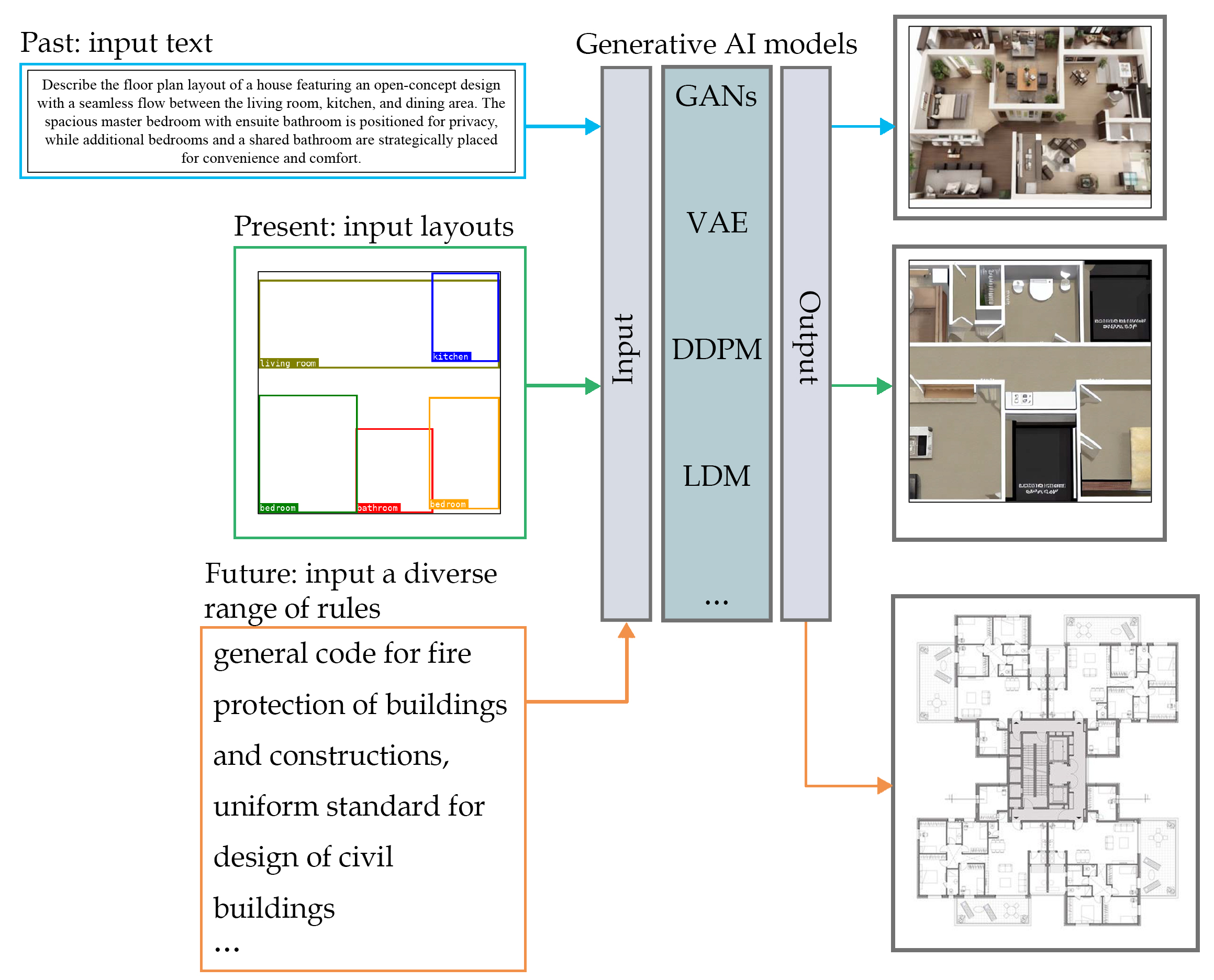}
    \caption{Existing generative models can generate layout of rooms based on input text and can also be controlled accordingly based on input layouts, Future models will be able to generate layouts using a diverse range of information.}
    \label{fig:4.1.1}
\end{figure}

The advancement of generative AI models offers innovative tools for generating and visualizing floor plans and spatial layouts through various inputs such as text and room layouts, as shown in Figure~\ref{fig:4.1.1}.
Text-to-image models like StackGAN \citep{zhang2017stackgan} and GLIDE \citep{nichol2021glide} can generate architectural floor plans from text prompts, enabling quick initial design drafts. Future improvements could integrate more inputs like building regulations, functional needs, and material details, producing detailed floor plans that comply with building standards in real-time for greater precision.
Image-to-image models like ControlNet \citep{zhang2023adding} can control outputs using layouts, sketches, and masks, while Layout2Im \citep{Zhao_2019_CVPR} generates architectural floor plans based on room layouts and spatial relationships. Future improvements could enhance detail accuracy and incorporate specific design parameters like wall and window placement, resulting in more refined, practical floor plans. Allowing real-time adjustments and generating multiple design options for comparison would make these models more efficient and adaptable design tools for various projects.
Video generation models like Make-A-Video \citep{singer2022make} and DynamiCrafter \citep{xing2023dynamicrafter} effectively display dynamic changes in architectural floor plans, aiding designers and clients in understanding spatial layouts. Future integration of real-time editing and interactive features could allow dynamic adjustments, such as modifying room layouts or spatial flow during video generation. This would create more intuitive and dynamic design presentations, enhancing communication efficiency in design presentations and client demonstrations.

\subsection{Future Research Directions for Architectural Facade Generation}

\begin{figure}[htbp]
    \centering
    \includegraphics[width=0.6\linewidth]{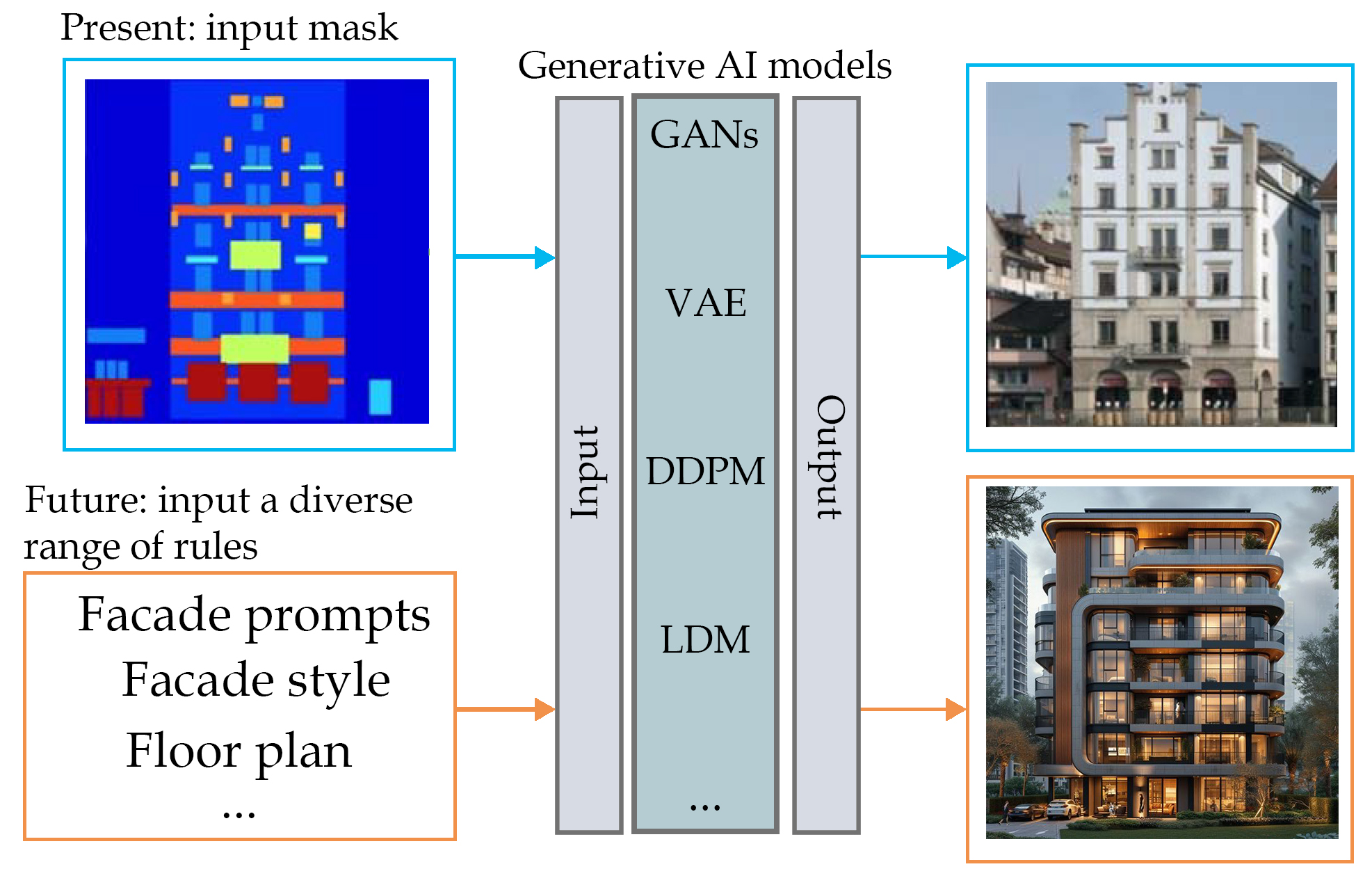}
    \caption{Current generative models can create facade based on input mask, Future models will be able to generate facade using a diverse range of information.}
    \label{fig:ex_4.1}
\end{figure}

Currently, layout and segmentation masks represent facade information in 2D image generation. Future applications of generative AI in facade generation may integrate various data inputs—such as semantic segmentation maps, conceptual images, and textual descriptors, as shown in Figure~\ref{fig:ex_4.1}.
Text-to-image models like GLIGEN \citep{li2023gligen} can generate architectural facade images from textual descriptions, quickly producing low-resolution sketches with basic shapes and colors, followed by refinement to enhance details and resolution. Future improvements could incorporate more input textual information like building regulations, material information, and style requirements to generate high-resolution facades that meet design standards. Adding real-time editing features would allow designers to adjust details flexibly during the generation process.
Similarly, image-to-image models like ControlNet \citep{zhang2023adding} generate aesthetically appealing facades based on images, prioritizing visual appeal over strict design standards. Future enhancements could include environmental adaptability (e.g., Heatmaps representing the light, ventilation, insulation) to ensure the facades are both visually pleasing and environmentally suitable.


\subsection{Future Research Directions for Architectural Structural System Generation}

Text/Image-to-image models like Midjourney \citep{zhong2024ai} and ControlNet \citep{zhang2023adding} inspire architectural design by generating creative structural images from text, layouts, or sketches. But they often overlook building codes, material properties, and mechanical requirements for complex structures. Future developments could focus on integrating building regulations and engineering constraints into the models, ensuring that generated designs are both creative and aligned with real-world standards.
Text-to-3D models like DreamFusion \citep{poole2023dreamfusion} generate 3D architectural structures, but the details often need to be more clear for complex designs. Future advancements will likely focus on improving precision and detail by refining algorithms and incorporating high-resolution 3D datasets. Integrating text-to-3D modeling with Building Information Modeling (BIM) could further enhance these models, enabling them to include not only geometric data but also technical parameters like material properties, energy consumption, and structural performance. AI-driven optimization could ensure that designs are both aesthetically pleasing and compliant with mechanical and safety standards.
Future image-to-3D models like CLIP-NeRF \citep{wang2021clip}, which combine text and image inputs, could enhance their ability to generate both stylized and functional outputs by incorporating users' style preferences, functional requirements, and technical constraints. Future developments may focus on leveraging deep learning to process complex multimodal inputs better.


\subsection{Future Research Directions for Architectural Section Generation}

Text/Image-to-image models like GLIDE \citep{nichol2021glide}  and Layout2Im \citep{Zhao_2019_CVPR} can transform text, hand-drawn sectional sketches, and layout diagrams into detailed sectional images. Future advancements could incorporate AI-driven optimization tools to improve precision, reducing the need for manual adjustments. Supporting multimodal inputs, such as text, sketches, and technical data, would enable the generation of sectional drawings that better align with project requirements. This evolution would greatly enhance the intuitiveness and efficiency of architectural design.
3D generation models like Magic3D \citep{Lin_2023_CVPR} construct 3D building models, allowing sectional images to be obtained by slicing, improving the understanding of complex spatial structures. Future advancements may integrate automated slicing and dynamic analysis tools, enabling architects to examine sections from different angles quickly. Enhanced precision will provide better insights into spatial layouts and structural details.
Video generation models like DynamiCrafter \citep{xing2023dynamicrafter} create dynamic visualizations of architectural sections, aiding in the comprehensive understanding of space performance. Future developments could integrate real-time simulation and environmental modeling, allowing designers to assess how building spaces perform under different conditions.

\section{Conclusion}

The following findings can be drawn from the research about generative AI models:
1) Advancements and Limitations of GANs: Generative Adversarial Networks (GANs) have laid a crucial foundation in the field of image generation. But issues such as mode collapse limit their application, necessitating future improvements in stability and diversity during generation. 
2) Impact of Diffusion Models: Diffusion Models (DMs), especially Latent Diffusion Models (LDMs), have significantly enhanced the quality and efficiency of image generation by reducing the dimensionality of the data, thereby ensuring high-fidelity output. 
3) Role of Foundation Models in Generative AI: Foundation models have demonstrated exceptional generalization capabilities in natural language processing (NLP) and computer vision, particularly in unsupervised and transfer learning, facilitating their broad application across multiple fields of generative AI. As generative AI models are increasingly applied to more complex tasks, such as video and 3D modelling generation, these areas are expected to become key directions for future research in generative AI models.

The following findings can be drawn from the research about generative AI for different architectural design steps:
1) Evolution of Generative AI Applications: generative AI applications in architectural design have evolved through three stages. Initially, the focus was on image generation. This progressed to creating images with 3D information. Currently, exploration includes generating diverse images, 3D models, and videos. This evolution has optimized the design steps, significantly improving efficiency in tasks such as conceptual image generation, 3D modelling, and floor plan design. Architects can now quickly generate a variety of creative designs.
2) Future Directions of Generative AI Applications: Advancements in technology are expected to enable generative AI to enhance personalization and real-time editing in architectural design. Specifically, developing Architectural 3D and image generative models will allow designers to make immediate adjustments based on user requirements. Future research may focus on integrating automatic performance optimization that considers building regulations and environmental standards, allowing designs to balance creative expression with functionality and sustainability.

This paper reviewed the advancements and applications of generative AI in architectural design. The following are the main contributions of this paper: 1) Provided a quick overview of generative AI technology development—this paper summarized the development of generative AI technologies, encompassing various visual and language generation models such as GANs, VAEs, visual large models, and LLMs. 2) Summarized the applications of generative AI in architectural design—this paper systematically organized and analyzed the application methods of generative AI in various architectural design tasks, including conceptual design, 3D form design, floor plan design, facade design, and structural design. A detailed review of relevant literature enables readers to quickly understand the current state of generative AI applications in the architectural design process. 3) Utilized case studies to illustrate the methods of applying AI in architecture—this paper demonstrated the innovative and practical value of generative AI in architectural design through specific application examples. It discussed in detail the different technology application scenarios in architectural design. These examples encompass various aspects such as buildings, urban blocks, and interior spaces, thereby providing readers with insights and inspiration for practical applications. 4) Predicted future directions and challenges—this paper proposed potential future research directions such as multidisciplinary integration, user-participatory interactive design, and the application of emerging generative AI models, thus providing reference points for further research. Moreover, this paper predicted the main challenges in the future application of generative AI in architectural design, including data acquisition and processing, model complexity, and professional barriers. Overall, this paper explored how integrating generative AI models into architectural design enhances schemes, simplifies processes, and boosts efficiency. 

Despite its contributions, this study has limitations that should be considered when interpreting the findings. The literature reviewed may not be fully comprehensive due to the random sampling of journals and conferences, potentially overlooking relevant studies. The research relies on existing literature, which may require updates as technology advances. The application of generative AI in section design is limited. Current AI technologies focus on specific design steps rather than the full architectural design process. Future research should explore more comprehensive AI applications across the entire design workflow.


\section*{Declaration of Competing Interest}

The authors declare that they have no known competing financial interests or personal relationships that could have appeared to influence the work reported in this paper.

\section*{Funding}
This work was supported by the Innovative Research Group Project of the National Natural Science Foundation of China [Award Number: 202401-202712].

\end{document}